\documentclass[11pt,a4paper]{article}
\usepackage[hyperref]{emnlp2020}
\usepackage{times}
\usepackage{latexsym}
\usepackage{graphicx}
\usepackage{amsmath}
\usepackage{booktabs}
\usepackage{enumitem}

\usepackage{microtype}

\aclfinalcopy 

\title{Dissecting Lottery Ticket Transformers: Structural and Behavioral Study of Sparse Neural Machine Translation}

\author{Rajiv Movva\thanks{\ \ Equal contribution.} \\
  MIT \\
  Cambridge, MA \\
  \texttt{rmovva@mit.edu} \\\And
  Jason Zhao\footnotemark[1] \\
  MIT \\
  Cambridge, MA \\
  \texttt{jzhao7@mit.edu} \\}
  
\date{}

\begin{document}
\maketitle
\begin{abstract}
Recent work on the lottery ticket hypothesis has produced highly sparse Transformers for NMT while maintaining BLEU. However, it is unclear how such pruning techniques affect a model's learned representations. By probing Transformers with more and more low-magnitude weights pruned away, we find that complex semantic information is first to be degraded. Analysis of internal activations reveals that higher layers diverge most over the course of pruning, gradually becoming less complex than their dense counterparts. Meanwhile, early layers of sparse models begin to perform more encoding. Attention mechanisms remain remarkably consistent as sparsity increases.
\end{abstract}

\section{Introduction}
In recent years, Transformers \citep{vaswani2017attention} have defined state-of-the-art performance on a variety of NLP tasks, including machine translation (MT) and language modeling. While large Transformer models can learn uniquely rich representations, they are also highly overparameterized \citep{michel_are_2019, hao_visualizing_2019}. Several studies have therefore attempted to prune Transformers during or after training while retaining as much performance as possible \cite{ganesh_compressing_2020}. Some methods have been fairly successful, achieving compression ratios up to 10$\times$ depending on the downstream task.

Looking beyond task performance, however, it remains unclear how widely-used pruning methods affect a model's learned representations. For example, a pruned Transformer may translate text at the same BLEU, but does pruning affect the model in ways unaccounted for by this metric?

Motivated by this question, we apply recent analysis techniques to study the representations of increasingly sparse Transformers trained on MT. We perform magnitude pruning in an iterative, lottery-ticket fashion to identify Transformers at competitive sparsities with no drop in task performance \citep{renda_comparing_2020, yu_playing_2020, brix-etal-2020-successfully}. We examine the internal structures of our models as sparsity increases, specifically addressing the following questions:

\begin{itemize}
\item Does pruning affect what linguistic knowledge is learned by the model?
\item How do individual model components (neurons, layers, attentions) change with pruning?
\item How is information distributed across layers in sparse vs. dense models?
\item What are the differences in pruning dynamics for the three types of model attention (encoder self, encoder-decoder, decoder self)? 
\end{itemize}

Using iterative magnitude pruning (IMP), we train an En-De Transformer that retains 99.4\% of BLEU at 66.4\% sparsity. During IMP, we obtain eight Transformer models at varying levels of sparsity, along with the original unpruned model. We probe these models' representations for learned linguistic knowledge on eighteen auxiliary syntactic and semantic tasks \citep{conneau2018cram, liu_linguistic_2019}. We then perform an unsupervised comparison of the representations and attention distributions between dense and sparse models, adopting metrics posed in \citet{wu_similarity_2020}. Our key conclusions are as follows:
\begin{itemize}
    \item Complex semantic information is lost first during pruning, before BLEU decreases.
    \item Model activations steadily diverge from their unpruned representations, particularly at higher layers.
    \item Information flow between layers becomes more distributed in sparse Transformers: lower layers perform more encoding.
    \item The encoder-decoder attention is the richest in representation, whereas the decoder self-attention is the simplest. Still, all attention mechanisms remain functionally consistent across sparsities.
\end{itemize}

\section{Related Work}

Much work has attempted to reduce the parameter count of dominant Transformer-based architectures \citep{ganesh_compressing_2020}. Several papers prune BERT \citep{devlin2018bert}, either via structured removal of layers and attention heads \citep{fan_reducing_2019, sajjad2020poor} or unstructured pruning of individual weights \citep{chen2020lottery, gordon2020compressing}. Structured head pruning has also been applied to NMT \citep{voita_analyzing_2019, michel_are_2019}, in which BLEU is used to quantify effective compression. Recent work from \citet{yu_playing_2020} uses iterative magnitude pruning to identify lottery tickets for NMT, retaining 99\% of BLEU at 67\% sparsity for Transformer-Big. To our knowledge, they achieve the highest net pruning ratio on translation with no drop in performance.

While most such studies are primarily considered with maximizing sparsity, a subset of them address other questions. \citet{gordon2020compressing} weight prune BERT and finetune on GLUE tasks to identify how much sparsity each task can accommodate. \citet{prasanna2020bert} prune heads while finetuning BERT on GLUE tasks, and identify which heads are masked most often. They use pruning as an analysis technique to identify `good' or `bad' BERT subnetworks. Similarly, \citet{michel_are_2019} and \citet{voita_analyzing_2019} prune heads to identify which types of attention are most relevant to performance. However, these studies focus only on task performance, leaving other behavioral differences between dense and sparse models unexplored.

Relevant methods of analyzing representations in NLP include probing classifiers, which evaluate model representations on supervised tasks for morphology \citep{Belinkov_2017}, syntax \citep{shi-etal-2016-string}, and/or semantics \citep{voita-etal-2018-context}. For Transformers, some work has directly examined the attention module \citep{Raganato2018AnAO, voita_analyzing_2019}. These analyses include inference of functional annotations for particular heads \citep{Clark2019WhatDB}, or assessment of attention's ability to perform unsupervised syntax tree prediction \citep{kim2020pretrained}. Recent work has also applied high-dimensional similarity analysis methods to compare learned representations within or across models \citep{saphra_understanding_2019, wu_similarity_2020}. For instance, \citet{bau2018identifying} identify recurring neurons across NMT models, interpret their functions, and control their activations. A broader survey of such literature is covered by \citet{belinkov2018analysis}. We leverage some of these representation analysis methods to study and compare sparse and dense Transformers, which, to our knowledge, previous work has not addressed.

\section{Generating Lottery Ticket Subnetworks}

\subsection{Training transformer-based NMT}

Following \citet{vaswani2017attention}, we train Transformer-Big on WMT16-En-De for 60 epochs and achieve detokenized test BLEU of 27.77 on Newstest14. Note that this score appears lower than Vaswani et al. since we do not use compound-split BLEU, which artificially inflates performance\footnote{When we compute compound-split BLEU with ensembling, our BLEU is 28.74, compromable with Vaswani et al.}.

\subsection{Iterative pruning \& rewinding protocol}
Recent work on the lottery ticket hypothesis has demonstrated the efficacy of iterative magnitude pruning (IMP) with \textit{weight rewinding} \citep{frankle2019stabilizing}, where unpruned network weights and the learning rate are rewound to values early in training after every pruning iteration. \citet{yu_playing_2020} apply this method to Transformers, and also show that leaving embedding weights unpruned better retains performance. \citet{renda_comparing_2020} propose \textit{learning rate (LR) rewinding}, where the learning rate is rewound to a value earlier in training, but the weights remain unchanged. They found that LR rewinding often performs better for deep NMT models and requires fewer training iterations.

Combining insights, we iteratively prune as follows: after training to completion, we mask the 20\% lowest magnitude non-embedding weights, rewind the LR to halfway through training (30 epochs), and retrain to completion before another prune. We also trained an iterative random pruning baseline using the same approach. As a clarifying note, we acknowledge that a ``lottery ticket'' traditionally refers to the network's pruned structure \textit{and} its weights early in training; however, for more convenient referral to our sparse models, we adopt a broader definition of the term to also describe network substructures identified via LR rewinding.

\subsection{Lottery ticket performance}

We iteratively pruned until our network's performance dropped significantly (Table \ref{tab:bleu}). After seven IMP steps, our model's non-embedding weights were 79\% sparse (sparsity including emb.~weights: 66.4\%), and test BLEU was over 99\% of its unpruned value (27.61 \& 27.77 respectively). In the subsequent pruning iteration, performance starts to drop more rapidly (0.4 BLEU with only 3\% of total additional weights being pruned), suggesting the start of the ``power-law'' performance decay observed during IMP \citep{rosenfeld_predictability_2020}. Our results align closely with \citet{yu_playing_2020}, who also report rapid BLEU drop at $>67\%$ net sparsity. Because we are primarily interested in sparse models that retain full performance, we stopped pruning at this iteration. For downstream experiments, we keep the seven pruned models which experience a negligible performance drop, as well as an eighth model to hint trends as pruning starts to degrade main task performance. In subsequent analyses, we refer to the model after the $k$th iteration of IMP as LTH$k$, with LTH0 referring to the unpruned model.

\subsection{Where are weights being pruned?}

Examining which model components are most readily pruned may hint at their relative importances. \citet{voita_analyzing_2019} find that late encoder-decoder heads and early decoder-decoder heads are retained the longest. We complement these findings with our results from unstructured pruning.

We compute sparsities of each weight module as overall Transformer sparsity increases (Figures \ref{fig:fcsparsity}, \ref{fig:attnsparsity}). For both the encoder and decoder, later layers exhibit higher fully connected sparsities, with as much as 25\% higher sparsity in decoder layer 6 FC weights compared to decoder layer 1 FC weights. This trend suggests that higher layers' FC modules are most overparameterized.     

For encoder and decoder self-attention (we compute sparsity across QKV weights $\mathbf{W_q,W_k,W_v}$ and out proj. matrix $\mathbf{W_o}$), layer 1 is pruned significantly more than other layers, particularly for low-sparsity models. Layer 6 is pruned next most. Meanwhile, encoder-decoder attention is pruned least across layers compared to all other modules: when overall model sparsity (excl.~embeddings) reaches 79\%, enc-dec sparsity is only 55\%. Late enc-dec layers are pruned slightly less than early ones. Finally, across all attention types, $\mathbf{W_v}$ and $\mathbf{W_o}$ are 25\% sparser than $\mathbf{W_q}$ and $\mathbf{W_k}$, suggesting that projection steps downstream of computing attention weights are particularly overparameterized.

\section{Probing for Linguistic Knowledge}

\subsection{Task setup}

Probing classifiers are used to measure latent linguistic knowledge in word embeddings \citep{belinkov_linguistic_2019}. We extract representations from our Transformer encoder and test whether they can be used to predict auxiliary labels about tokens or pairs of tokens from external datasets.

\citet{liu_linguistic_2019} release a suite of probing tasks of varying linguistic complexity, and we largely inherit this setup to study our pruned networks. For streamlined analysis, we broadly split the eighteen tasks into three groups as follows (some tasks may span multiple categories): \textit{Part-of-Speech:} Penn TreeBank POS tagging \textbf{(POS)}; \textit{Syntactic:} CCG supertagging \textbf{(CCG)}, parent, grandparent, and greatgrandparent ancestor prediction \textbf{(Parent, GParent, GGParent)}, chunking \textbf{(Chunk)}, named entity recognition \textbf{(NER)}, grammatical error detection \textbf{(GED)}, conjunct identification \textbf{(Conj)}, \textbf{syntactic arc prediction}, and \textbf{syntactic arc classification}; \textit{Semantic:} semantic tagging \textbf{(ST)}, preposition lexical function and semantic role disambiguation \textbf{(PS-Fxn, PS-Role)}, event factuality \textbf{(EF)}, \textbf{semantic arc prediction}, \textbf{semantic arc classification}, and coreference arc prediction \textbf{(Coref)}. See \citet{liu_linguistic_2019} for detailed task descriptions.

\paragraph{Implementation details.} We use the same data, splits, and evaluation as \citet{liu_linguistic_2019}. Our initial probing experiments use a single linear layer mapping our 1024-dim token embeddings to the number of task outputs. We train separate probes for each of the six encoder layers for the first nine LTH iterations (LTH0 unpruned, LTH1--8 pruned). See Appendix 1 for further implementation notes.

\subsection{Initial results}

\begin{figure*}[!htb]
    \centering
    \includegraphics[width=\textwidth]{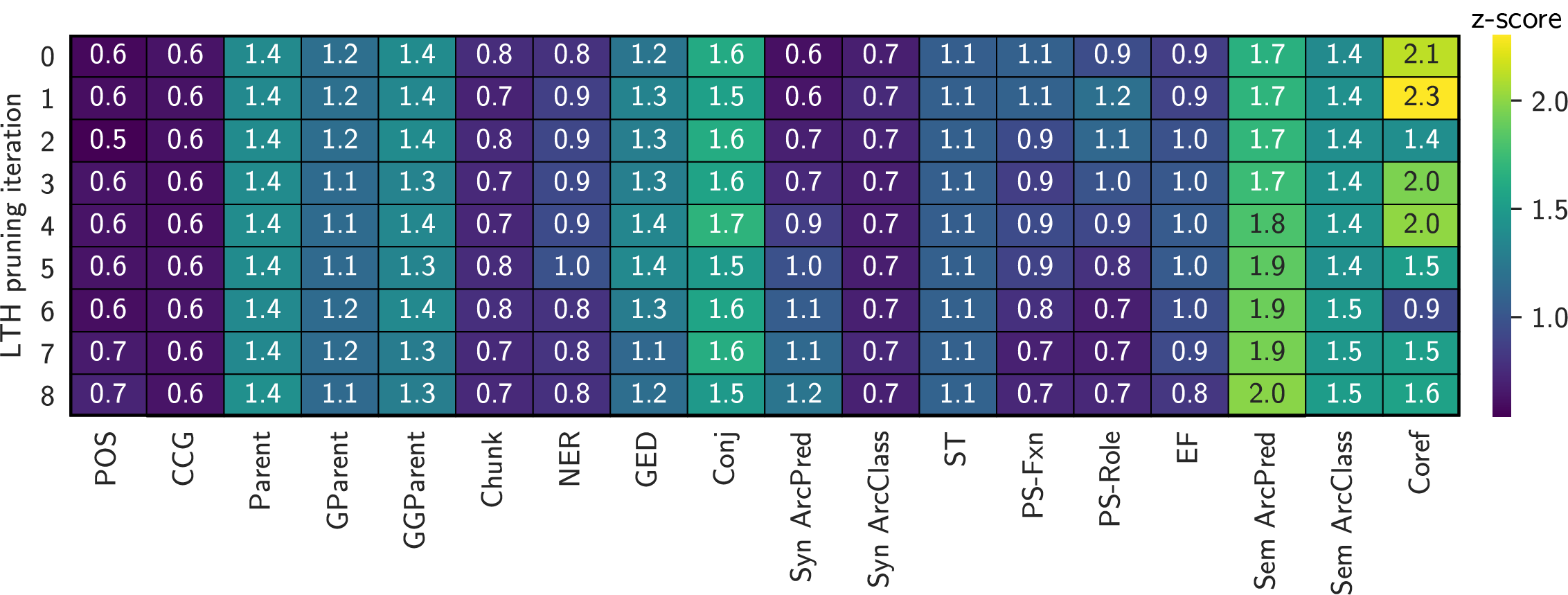}
    \caption{Each cell shows a model's best linear probing performance across all encoder layers for a particular task. Sparsity increases from top to bottom. Values shown are task-specific $z$-scores. }
    \label{fig:probingtable}
\end{figure*}

In Figure \ref{fig:probingtable}, we show each LTH model's best probing performance across all layers for each task. Several tasks of varying complexity -- POS, CCG, ancestor prediction, syntactic chunking, conjunct ID, syntactic arc classification, semantic tagging, and semantic arc classification -- are \textit{sparsity-invariant}, \textit{i.e.}~probing performance does not exhibit any sparsity-specific trend. Our metrics for these tasks are on par with probing classifiers trained on \textit{e.g.} BERT contextual word embeddings and are consistently higher than results from a GLoVe baseline \citep{liu_linguistic_2019}, suggesting that our probes effectively learn these tasks at all sparsities. We conclude that sparsity-invariant tasks encode necessary information for NMT, so their performance will be maintained as long as BLEU remains high enough. Relatedly, we note that some of these tasks (e.g. POS, CCG, ST) have near-100\% accuracy and smaller relative improvements over baselines \citep{liu_linguistic_2019}, so they may only require `simpler' types of linguistic knowledge contained in any competent language encoder.

Next, there are \textit{sparsity-degrading} tasks, for which the probe does best at an early pruning iteration and starts to drop off at higher sparsities. These tasks include PS-Fxn, PS-Role, and Coref, in which best performance is achieved at LTH0 or LTH1, and performance starts to drop at LTH5 (56.5\% sparsity). PS-Fxn classifies a preposition's lexical function in its prepositional phrase, while PS-Role identifies the semantic role that the preposition confers to its object \citep{schneider_comprehensive_2018}. Both tasks require integration of semantic knowledge on top of standard syntax parsing; baseline GLoVe performance is very weak compared to deep contextual representations \citep{liu_linguistic_2019}.  While preposition disambiguation is important to sentence understanding, STREUSLE annotations are likely more fine-grained than necessary for correct translations; even early NMT models could accurately translate most prepositions \cite{isabelle_challenge_2017}. As a result, sparse models could plausibly lose some information relevant to this probing task without impacting BLEU.

Meanwhile, coreference resolution involves identifying pairs of words referring to the same object. NMT models struggle with coreference when semantic information contradicts stereotypical patterns in the training set, \textit{e.g.} for gendered pronouns \citep{ stanovsky_evaluating_2019}. However, these cases are generally rare and investigated with specific challenge sets, and they may not manifest noticeably in test BLEU.

Interestingly, we found that syntactic and semantic arc prediction were \textit{sparsity-improving}; sparser networks consistently performed better. Syntactic arc prediction aims to identify links between co-dependent words in a parse tree, while semantic arc prediction links objects related by the question \textit{Who did What to Whom?} \citep{oepen_semeval_2015}. Both tasks are difficult for GLoVe embeddings, but MT-derived representations have done well \citep{liu_linguistic_2019, belinkov_linguistic_2019}. Performance on these arc prediction tasks is lower than on their arc classification counterparts, which were both sparsity-invariant. 

Summarizing, we conclude that (1) sparsity-invariant tasks represent core linguistic information that remains encoded as long as BLEU is high enough; (2) there is a push-and-pull with higher-order features as sparsity increases, with some knowledge becoming more readily-available to a probe as other knowledge is degraded. 

\subsection{Are results probe-sensitive?} 

\begin{figure}[!htb]
    \centering
    \hspace*{-0.2in}
    \includegraphics[width=0.45\textwidth]{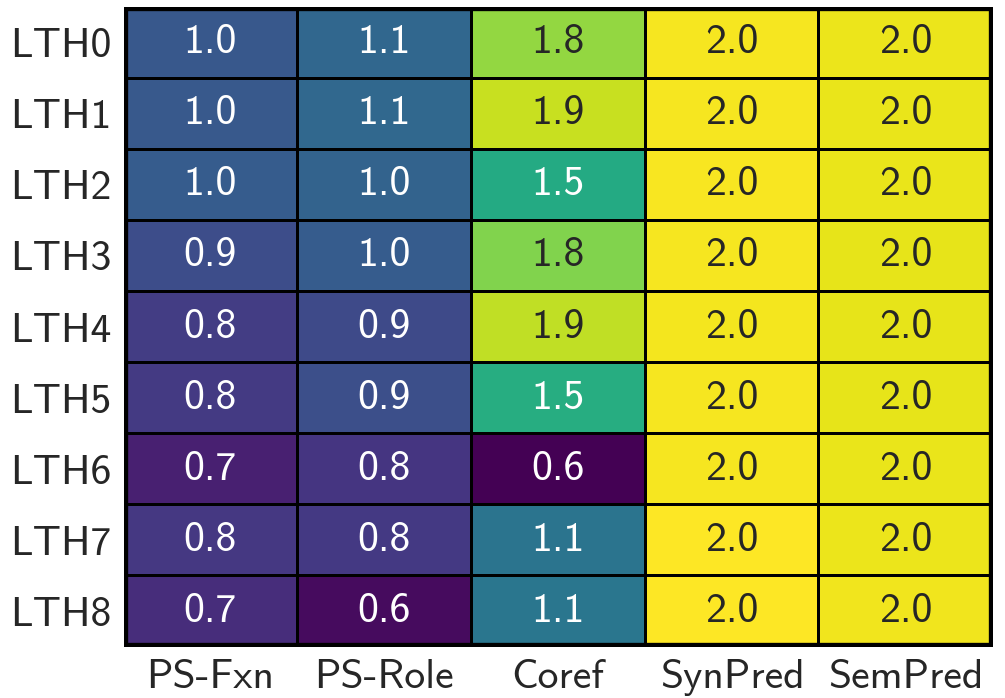}
    \caption{Each model's best performance using the MLP probe, for five tasks whose linear probe performance varied with sparsity. We report $z$-scores.}
    \label{fig:mlpresults}
\end{figure}

Minimal probes offer efficient comparison of language encoders \cite{hewitt_designing_2019}, but more complex probes can also provide useful results when taken in context \cite{pimentel_information-theoretic_2020}. We wondered if sparsity-specific performance differences would hold up to a more complex probe, so we repeated a subset of tasks using a two-layer multilayer perceptron. This probe family had over an order of magnitude more parameters than the linear probes. We show performance $z$-scores in Figure \ref{fig:mlpresults}, with raw performances available in Table \ref{tab:mlpresults}.

Across the board, raw accuracies were higher using the MLP. We no longer see a performance improvement in sparse models for syntactic and semantic arc prediction. Thus, dense model encodings contain the necessary information for these tasks, but a probe must have enough weights to extract it. Given that the Transformer decoder has far more parameters than either probe, it is not surprising that less direct representations of some linguistic features would not impact translation performance. Differences in how directly sparse \& dense models encode information is an interesting question perhaps well-suited to recent work on minimum description lengths \citep{voita_information-theoretic_2020}, but we leave it to future study. Meanwhile, the MLP could not rescue sparse model performance on PS-Fxn, PS-Role, and Coref; results were nearly identical as with the linear probe. We conclude that pruning corrupts some semantic knowledge relevant to these three probing tasks. 

\subsection{Layer-specific trends}

\begin{figure}[!htb]
    \centering
    \includegraphics[width=0.5\textwidth]{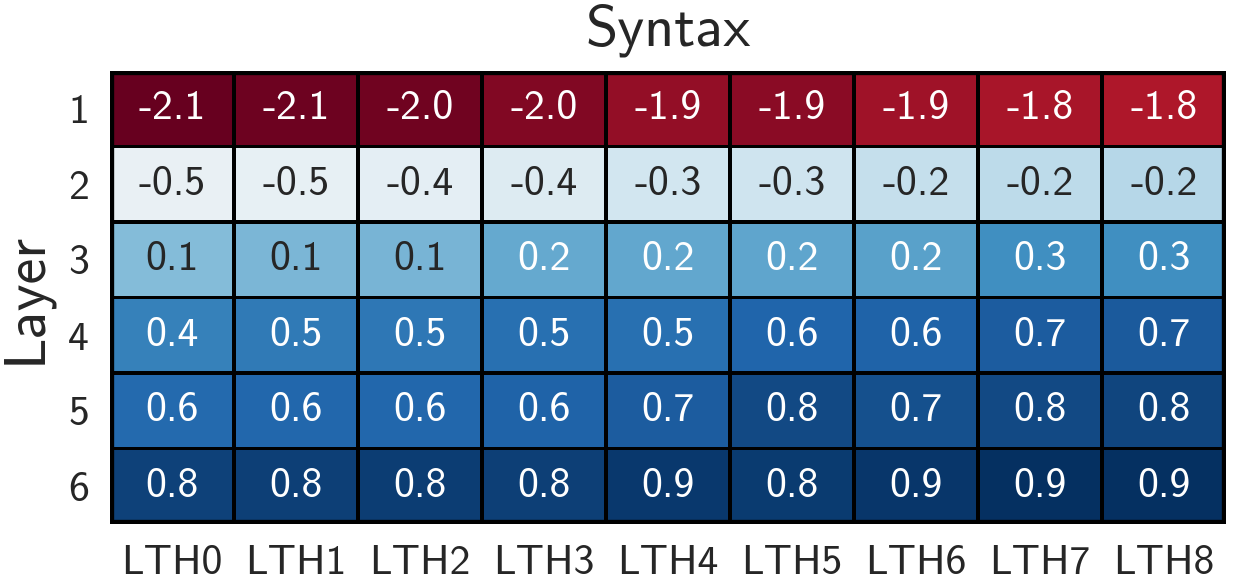}
    \hspace*{-0.2in}
    
    
    \includegraphics[width=0.5\textwidth]{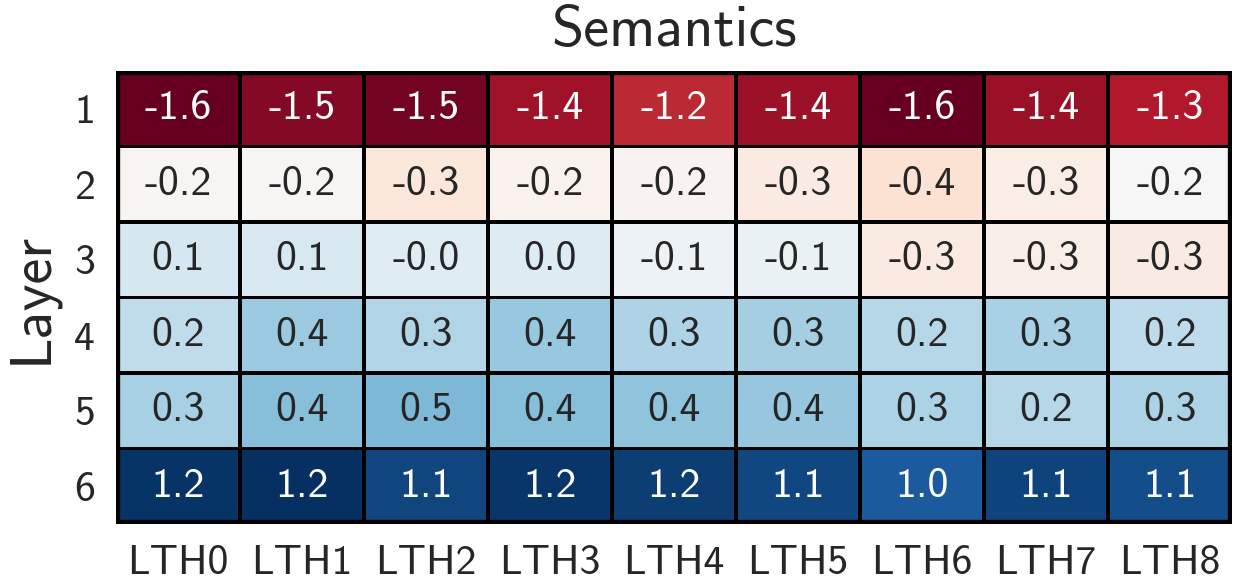}
    \hspace*{-0.2in}
    \caption{Each cell shows the average probing $z$-score (across all syntactic or semantic tasks) for a particular layer of a particular model. Model sparsity increases from left to right.}
    \label{fig:probinggroups}
\end{figure}

Our analysis so far has focused only on probing performance at the final layer (which always had highest accuracy); we next wanted to study any potential layer-specific sparsity trends. In Figure \ref{fig:probinggroups}, we show average linear probe performance for each layer of each model with tasks grouped as syntactic or semantic. For syntactic tasks, performance using layer 1-5 representations increases with sparsity, suggesting that lower layers of sparse models better learn syntactic information. However, performance is maximized and equal across sparsities by layer 6. Results on POS tagging show the same trend (Figure \ref{fig:posprobing}). For semantic tasks, all models perform similarly at early layers, while dense models slightly outperform by the final layers. These results support an interpretation in which early sparse layers more directly encode low-level information, whereas dense models tend to rely more on their final layers to (1) equalize differences on syntax tasks and (2) outperform on semantic tasks.

\section{Measuring Behavioral Similarities of Sparse \& Dense Models}

\subsection{Experimental setup}

Probing classifiers are one method of studying linguistic knowledge, but even two models with similar probing results may have divergent representations \cite{saphra_understanding_2019}. We are left with the question: do sparse models learn to arrive at the same internal representations as dense models using fewer weights (up to linear transformation), or do their activations and attention maps shift altogether? To answer this question, we perform direct, unsupervised study of our model's internal vectors on unseen text.

First, we deploy our unpruned model and the eight pruned models on our 3000 validation sentences, which span 120K tokens total. We store the 1024-dim token representations and the 16 heads of each of the three attention types (enc-enc, enc-dec, dec-dec) at each layer. To compute similarities from these data, we adopt a subset of the metrics described in \citet{wu_similarity_2020}. Each metric offers a distinct lens of viewing the behavioral similarity of a layer pair $(L, L')$.

\textbf{NeuronSim} \citep{bau2018identifying} is a local similarity measure that quantifies how well the individual neurons $k$ of a layer $L$ (\textit{i.e.}, a single dimension of a layer's representation) align with individual neurons $k'$ in another layer $L'$: 
$$\text{NeuronSim}(L, L') = \underset{k\in L}{\mathrm{mean}} \left\{\max_{k' \in L'} \mathrm{corr}(k, k')\right\},$$ where $\mathrm{corr}(k, k')$ is the mean Pearson correlation between the activations of neurons $k$ and $k'$ across all tokens. \textbf{LayerSim} is a global similarity measure that quantifies how well entire layer representations align. For two layers, LayerSim compares their vectors of 1024-dim representations across all tokens. We use the linear centered kernel alignment (linearCKA) \citep{kornblith_similarity_2019} as our similarity metric. Analogous to LayerSim, \textbf{AttentionSim} compares the attention distributions across all heads between two different layers. For every sentence $s$ in our dataset, we extract the 16 head attention vector $\boldsymbol{\alpha}_{ij}(s)$ for each word pair $(w_i, w_j) \in s$. We then apply linearCKA, treating each pair of words as a 16-dimensional example. 

\subsection{Similarity of activations}

\begin{figure}[!ht]
    \centering
    \includegraphics[width=0.48\textwidth]{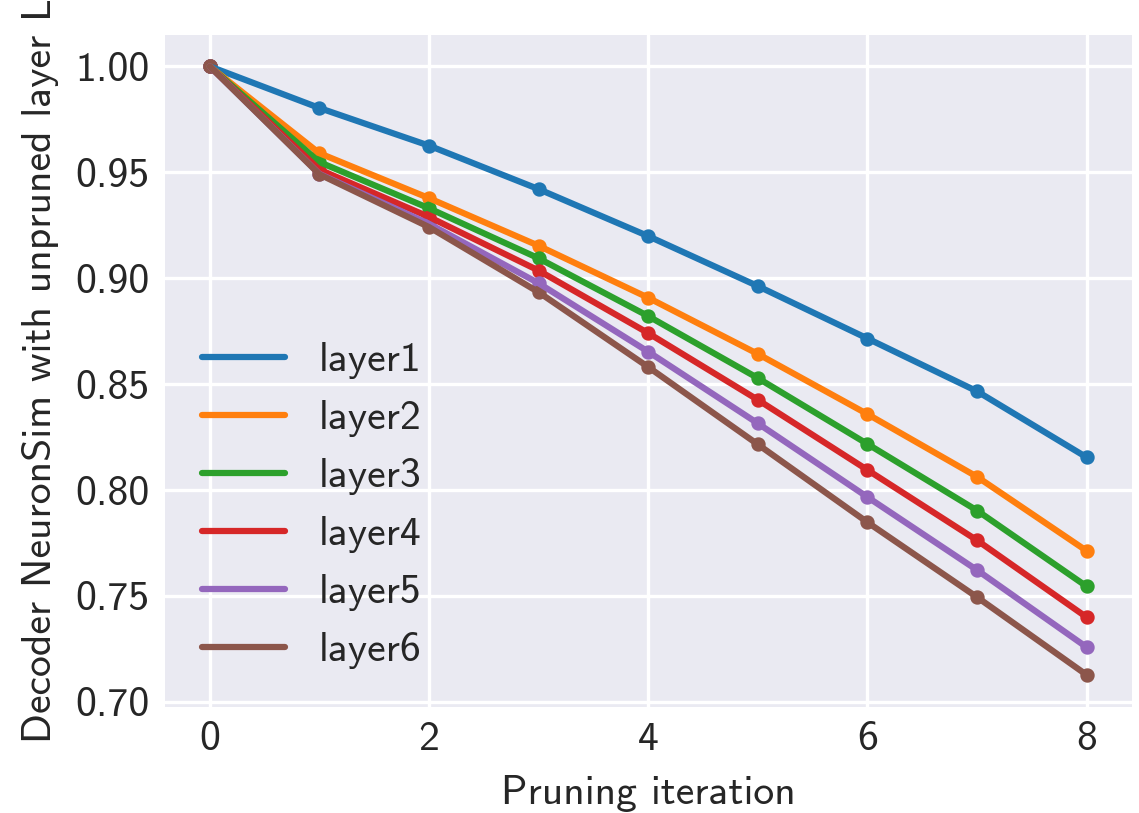}
    \caption{Each decoder layer's NeuronSim with the corresponding LTH0 layer; sparsity increases left to right.}
    \label{fig:neuronsimlineplot}
\end{figure}

Looking first at NeuronSim, we find that neuron function in the encoder and decoder is largely conserved as sparsity increases. Across all neurons $k$ in layer $L$ of LTH0 (unpruned), $k$'s most correlated neuron in layer $L$ of LTH8 (70\% sparse) is the same neuron $k$, 99.8\% of the time. However, the \textit{magnitude} of neuron similarity with the unpruned network consistently drops with sparsity (Figure \ref{fig:neuronsimlineplot}). In both the encoder and decoder, this drop was sharpest for higher layers: \textit{e.g.} between LTH0 and LTH8, decoder layer 1 had 0.82 NeuronSim while decoder layer 6 had 0.71 NeuronSim.

Next, we wondered if all neurons gradually become less similar to their unpruned selves, or rather if some neurons remain the same whereas others ``drop out'' or change functions entirely. Visualizing the distributions of neuron correlations revealed the former: all sparse-dense neuron pairs became less similar during pruning (Figure \ref{fig:maxcorrdists}). We therefore conclude that as sparsity increases, (1) neurons gradually diverge from their dense counterparts, and (2) neurons in higher layers diverge more rapidly than neurons in lower layers.

\begin{figure}[!htb]
    \centering
    \hspace*{-0.2in}
    \includegraphics[width=0.5\textwidth]{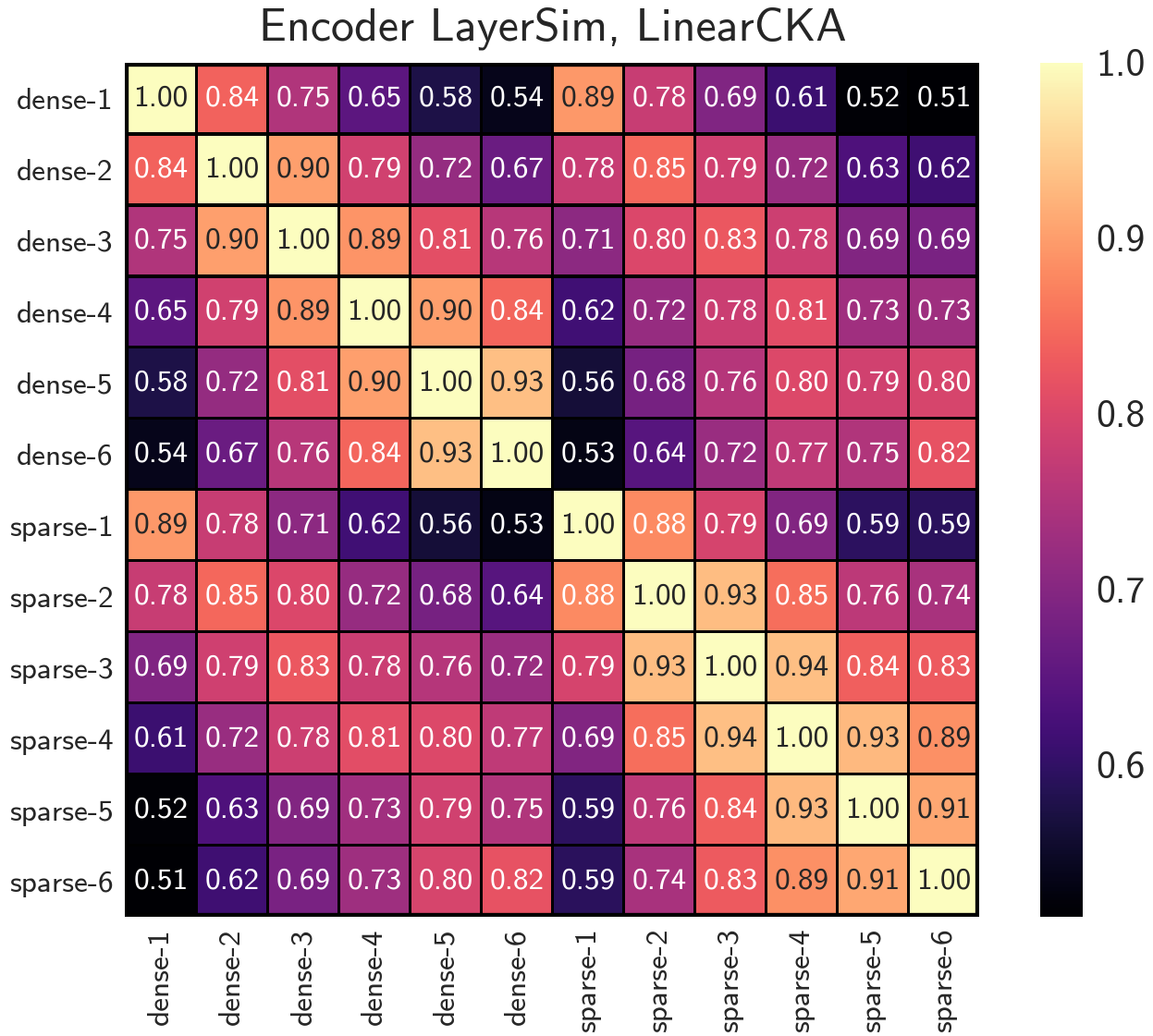}
    \caption{Encoder layer similarities for pairs of layers in LTH0 (dense) and LTH8 (70\% sparse).}
    \label{fig:encoder_layersim}
\end{figure}

In Figure \ref{fig:encoder_layersim}, we compute encoder LayerSim scores between LTH0 (dense) and LTH8 (sparse). Like NeuronSim, we find that similarity decreases with sparsity, especially at higher layers. Interestingly, in the decoder, LayerSim between dense and sparse was consistently higher at layer 6 than layer 5 (Figure \ref{fig:decoder_layersim_lineplot}), perhaps because layer 6 representations `converge' before final token prediction. Comparing off-diagonal layer similarities between encoder and decoder, we find that different decoder layers are less similar than different encoder layers (Figures \ref{fig:encoder_layersim}, \ref{fig:decoder_layersim}). That is, each decoder layer changes token representations more significantly than each encoder layer, suggesting that decoder layers perform more processing than encoder layers (perhaps due to the additional parameterization afforded by the encoder-decoder attention module).

Next, we find that early sparse model representations are closer to their final representations than early dense representations are. In the encoder, for example, sim(dense-2, dense-6) is 0.67, while sim(sparse-2, sparse-6) is 0.74 (Figure \ref{fig:encoder_layersim}). In the decoder, sim(dense-4, dense-6) is 0.80 while sim(sparse-4, sparse-6) is 0.92. In general, early and late layers are more similar in sparse models, and this trend strengthens as sparsity increases.

We hypothesized two explanations: (1) Sparse models have less complex final representations, so early layers are inherently not as `far away.' (2) Early sparse layers learn more effective representations, so they are closer to the final ones. Looking at decoder LayerSim values, we found that early sparse representations were often more similar to final \textit{dense} representations than early dense representations were, providing some evidence for (2). For example, sim(sparse-2, dense-6) is 0.55 while sim(dense-2, dense-6) is 0.49 (Figure \ref{fig:decoder_layersim}).

\subsection{Analyzing the final layers}

Late encoder and decoder representations exhibited largest difference between sparse and dense models. To further characterize the difference, we computed the SVD of each model's final encoder and decoder layer representation matrices. We find that sparse models have more variance explained by the top $k$ singular vectors (Figure \ref{fig:svd}). For example, for the encoder, 80\% variance requires $k=290$ for LTH0 but just $k=176$ for LTH8; for the decoder, $k=283$ and $k=139$. We conclude that final layer representations in sparse models fundamentally have less mathematical complexity (which is not necessarily an obvious result for magnitude pruning, versus \textit{e.g.}~in the case of pruning entire neurons).

We next found word categories for which dense and sparse encoder representations differed most. We used linearCKA to compute similarities of tokens grouped by frequency bin, POS (Penn TreeBank), or semantic tag \citep{bjerva_semantic_2016}.

Mid-frequency tokens (rank 100 to 500) had highest similarity (0.95), while both common (rank 0 to 5) and rare (rank 2500 or higher) tokens had similarity 0.87. For POS, coordinating conjunctions and superlative adjectives had highest similarity (0.98 and 0.97), while proper nouns and particles were lowest (0.85, 0.86). Models also learned very different representations of punctuation, \textit{e.g.}~with the possessive ending and period tokens at 0.79 and 0.84 similarity respectively. For semantic tags, the least similar classes (0.84) were perfect/progressive verb tense modifiers, \textit{e.g.}~`has arrived', `is running', etc. The broad `concept' class spanning uncommon nouns also had relatively dissimilar (0.87) representations. Together, these results suggest that sparse model encodings differ most for (1) tokens with several syntactic/semantic meanings to disambiguate and (2) rare words.

\subsection{Attention-level similarities}

\subsubsection{Encoder self-attention}

\begin{figure}[!htb]
    \centering
    \hspace*{-0.2in}
    \includegraphics[width=0.5\textwidth]{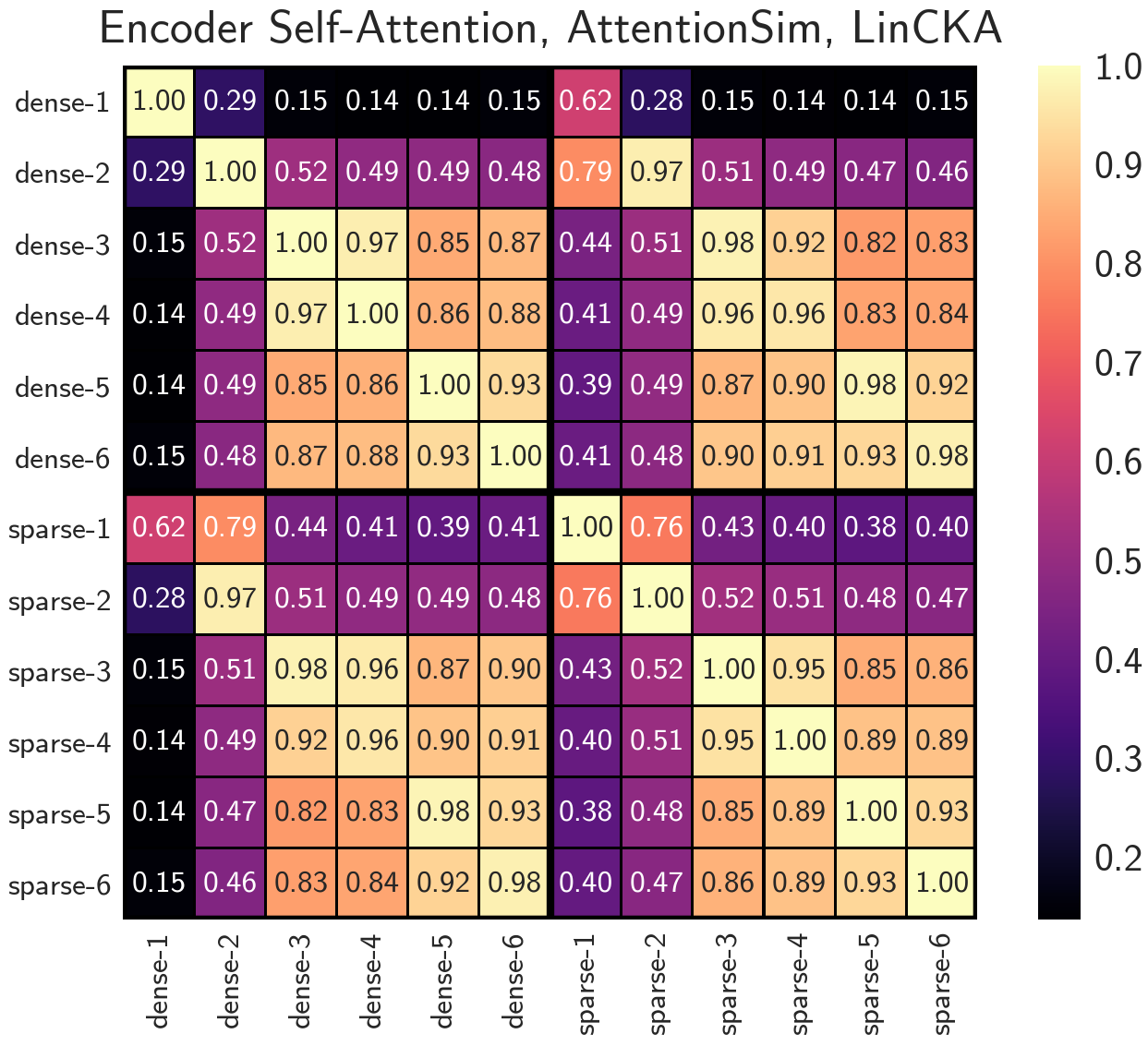}
    \caption{Encoder self-attention similarities for pairs of layers in LTH0 (dense) and LTH8 (70\% sparse).}
    \label{fig:encenc-attentionsim}
\end{figure}

Unlike the encoder activations, encoder self-attention distributions remain remarkably similar between sparse and dense models. For layers 2--6, AttentionSim scores on the sparse-dense diagonal are very close to 1 (Figure \ref{fig:encenc-attentionsim}, bottom left). Layer 1 is an anomaly, with sparse and dense attentions differing widely (0.62). The first layer's attention distributions in the sparse model become much more similar to its later layers (average 0.41) than in the dense model (average 0.18), suggesting that this first self-attention layer learns more salient relationships in the sparse model.

\subsubsection{Encoder-decoder attention}

Compared to self-attention, encoder-decoder attention displays more variation across layers and sparsities (Figure \ref{fig:attentionsim_small}). While many off-diagonal similarities exceed 0.85 in self-attention, off-diagonal encoder-decoder similarities are often less than 0.7. In particular, the first three enc-dec layers differ strongly from the last three layers, demonstrating heterogeneity in learned attention distributions at different levels of decoding.

As sparsity increases, attention maps at a given layer remain mostly consistent with the dense model, although there is slightly more deviation than in self-attention. Interestingly, the model's off-diagonal similarities gradually increase with sparsity (average 0.69 in LTH8 vs. 0.63 in LTH0), particularly with sparse-5 becoming more similar to sparse-1,2,3. Sparsity may have a dampening effect on the distinctions of individual encoder-decoder layers; these more homogenous attention distributions may explain the drop in the decoder's total representational complexity (Figure \ref{fig:svd}).

\subsubsection{Decoder self-attention}

Decoder self-attention distributions in pruned models are almost identical (0.99 similarity) to their corresponding unpruned distributions, even more so than enc-self and enc-dec attention (Figure \ref{fig:attentionsim_small}). We attribute this phenomenon to the relative simplicity of the decoder self-attention module, and in consequence the relative ease at which weights can be pruned without changing expressivity. For instance, we found that 40\% of all decoder self-attention distributions in the sparse model (41\% in the dense model) placed over 0.95 of the probability mass on a single query token, compared to only 30\% and 16\% of the distributions in enc-self and enc-dec attention respectively.

Further, all similarity scores between different layers (off-diagonal) in dec self-attention are significantly higher than in encoder self and enc-dec attention. That is, decoder self-attention is homogenous across layers. Despite the simple nature of this self-attention, the decoder can still learn complex representations due to its pairing with the powerful encoder-decoder attention module.

\section{Discussion}

A consistent theme in our analysis is the behavioral shift of early layers (1--3), which occurs gradually as sparsity increases. Our probing results find that lower layers of sparse models more directly encode POS and syntax information compared to dense models, even though performance of the final encoder representations is similar (4.4). Moreover, our similarity analyses conclude that early layer encoder hidden representations (5.2) and attention distributions (5.4.1) trend closer towards their respective final representations in sparse models. Information-theoretically, sparse layers have less maximum capacity for encoding, so each individual layer must shoulder more load for the final representations to remain predictively salient. Conversely, an overparameterized dense model can compensate for weak lower layer representations with its upper layers. Indeed, upper FC layers are pruned more than lower FC layers (3.4), reflecting the shift in modeling power away from higher layers.

We also observe a gradual loss of information stored in model representations as weights are pruned, especially in later layers. Individual neurons diverge from their dense counterparts (5.2), causing a drop in overall representational complexity in the encoder and decoder. Correspondingly, sparse models perform worse at higher-order semantic tasks that are less relevant to BLEU (4.3). The reduced overall complexity of sparse representations may partially explain why final layers are observed to be closer to early layers (5.2, 5.4.2).

Finally, we find that sparse models' attention distributions remain largely similar to their values in the dense model. This ability to reduce weights in attention modules while maintaining nearly identical representations affirms other lines of work \citep{guo2019startransformer, wang2020linformer}. Of the three attention types, encoder-decoder is pruned least (3.4), varies most across sparsities, and exhibits most within-model, inter-layer heterogeneity (5.4.3). These results corroborate existing evidence of its unique importance \cite{voita_analyzing_2019, michel_are_2019}. Meanwhile, decoder self-attention is extremely homogenous across layers and sparsities, perhaps because encoder-decoder attention is more relevant to creating rich representations.

\paragraph{Limitations.} Our work focuses on pruned Transformers for which BLEU remains similar to the original model. However, BLEU is an imperfect measure of translation quality \citep{callison-burch-etal-2006-evaluating}, and it is possible that our pruned models actually perform worse on the task at lower sparsities than suggested by BLEU. Still, we think our work is relevant given that sparse models are typically only held to the standard of matching unpruned task performance.

Next, we emphasize that our work focuses solely on magnitude pruning, which may not be representative of how other pruning methods impact Transformers. We chose this style of pruning primarily because it allows for higher overall sparsity without drop in performance \cite{renda_comparing_2020}. Further, while it might be expected (and has been shown, in some cases) that pruning entire neurons or attention heads would substantially change \textit{e.g.}~the distributions of the model's outputs, we found less existing work specifically measuring the effects of magnitude pruning. This dearth of analysis seemed particularly egregious given recent growth in work on unstructured sparsity \citep{blalock_what_2020}.

Finally, a note on probing classifiers: as has been widely discussed by the community (\textit{e.g.}~\citet{pimentel_information-theoretic_2020}), probes measure \textit{correlation} between model outputs and auxilliary information. Differences in probe performance do not necessarily imply anything about what information actually uses during its forward pass. Especially since we find some evidence suggesting that sparse models may be encoding information across layers, it is possible that their differing structure may explain worse probe performance, as opposed to fundamentally weaker linguistic feature extraction. We hope future work supplements our results by analyzing a model's encoded knowledge in other ways.

\section{Conclusions}

We evaluate how unstructured pruning affects the behavior of Transformers while task performance is maintained. We use probing classifiers to demonstrate that pruning degrades semantic knowledge before affecting BLEU, and that early layers of sparse models better encode low-level linguistic information. Unsupervised similarity analysis reveals that pruning induces representational changes in the encoder and decoder, particularly in higher layers, and that early sparse representations are more similar to their final representations. Meanwhile, attention distributions remain remarkably similar, even at high sparsities.

\section*{Acknowledgements}

We thank Yonatan Belinkov and Jonathan Frankle for their advice during the initial stages of the project. We thank Nelson F. Liu for providing access to preprocessed probing datasets.

\bibliographystyle{acl_natbib}
\bibliography{anthology,nlp-pruning-interpretability}

\clearpage

\appendix

\renewcommand\thefigure{\thesection\arabic{figure}}
\setcounter{figure}{0}

\renewcommand\thetable{\thesection\arabic{table}}
\setcounter{table}{0}

\section{Appendix}

\subsection{Probing Implementation Details}

For initial experiments, we trained a linear probe mapping our 1024-dimension token representations to the number of output classes (e.g. 45 POS tags). For subsequent MLP probing, we use an MLP with one 1024-neuron hidden layer with ReLU activation. All weights are trained using Adam with learning rate $10^{-3}$ for at most 50 epochs with 3 epochs of early stopping patience. 

For more complete task descriptions, please refer to \citet{liu_linguistic_2019}. Of the eighteen tasks, five are pairwise, \textit{i.e.} they involve predicting a property about a pair of tokens. These tasks are syntactic arc prediction, syntactic arc classification, semantic arc prediction, semantic arc classification, and coreference resolution (note that prediction refers to binary link identification while classification concerns the type of link). For these prediction tasks involving pairs of tokens, we input the two token embeddings $\mathbf{w}_i$ and $\mathbf{w}_j$ in addition to their elementwise product $\mathbf{w}_i \odot \mathbf{w}_j$ (as in \citet{liu_linguistic_2019}). 

Because our model uses Moses tokenization and byte-pair encoding, the source tokens in our preprocessed probing datasets are further split into subtokens by our model. We aggregate subtoken representations by averaging representations. Finally, we noticed that tasks with smaller train/test sets displayed some run-to-run variability, so for these tasks (PS-Fxn, PS-Role, Coref), we report averaged metrics across five replicate runs with different random seeds (for both the linear probe and for the MLP probe).

\begin{table*}
\centering
\caption{Sparsities and BLEUs for the pruned models at each pruning iteration. LTH$k$ refers to the name of the model pruned using learning rate rewinding, after $k$ pruning iterations. Because we don't prune embedding weights, we computed two sparsity values, one including all weights and the other excluding the embedding weights. MP BLEU refers to the models pruned using magnitude pruning (\textit{i.e.} the models we perform analysis of), while Random BLEU refers to the baseline using iterative \textit{random} pruning with LR rewinding.}\label{tab:bleu}
\begin{tabular}{@{}ccccc@{}}
\toprule
  \multicolumn{1}{c}{\begin{tabular}[c]{@{}c@{}}Model\\ Name\end{tabular}} &
  \multicolumn{1}{c}{\begin{tabular}[c]{@{}c@{}}Sparsity\\ (incl. emb)\end{tabular}} &
  \multicolumn{1}{c}{\begin{tabular}[c]{@{}c@{}}Sparsity\\ (excl. emb)\end{tabular}} &
  \multicolumn{1}{c}{\begin{tabular}[c]{@{}c@{}}MP \\ BLEU\end{tabular}} &
  \multicolumn{1}{c}{\begin{tabular}[c]{@{}c@{}}Random\\ BLEU\end{tabular}} \\ \midrule
LTH0 & 0.000 & 0.000 & 27.77 & 27.77 \\
LTH1 & 0.168 & 0.200 & 28.04 & 27.59 \\
LTH2 & 0.302 & 0.360 & 28.00 & 27.81 \\ 
LTH3 & 0.410 & 0.488 & 27.70 & 27.46 \\
LTH4 & 0.496 & 0.590 & 27.93 & 27.24 \\
LTH5 & 0.565 & 0.672 & 27.80 & 26.90 \\
LTH6 & 0.620 & 0.738 & 27.76 & 26.51 \\
LTH7 & 0.664 & 0.790 & 27.61 & 26.14 \\ 
LTH8 & 0.669 & 0.832 & 27.19 & 25.82 \\
LTH9 & 0.727 & 0.865 & 27.16 & 25.33 \\
\bottomrule
\end{tabular}
\end{table*}

\begin{table*}
\centering
\caption{Results using the linear probe, for the subset of tasks whose performances vary with sparsity.}
\begin{tabular}{lcccccccc}
\toprule
\bf{Model} &  PS-Fxn &  PS-Role &  Coref &  SynPred &  SemPred &   NER &   GED &    EF \\ 
\midrule LTH0 &   0.858 &    0.740 &  0.771 &        0.905 &        0.892 & 0.718 & 0.302 & 0.712 \\ 
LTH1 &   0.859 &    0.760 &  0.778 &        0.906 &        0.894 & 0.723 & 0.307 & 0.714 \\ 
LTH2 &   0.847 &    0.756 &  0.743 &        0.908 &        0.895 & 0.726 & 0.305 & 0.719 \\ 
LTH3 &   0.840 &    0.744 &  0.764 &        0.909 &        0.898 & 0.727 & 0.306 & 0.722 \\ 
LTH4 &   0.847 &    0.737 &  0.766 &        0.912 &        0.903 & 0.728 & 0.309 & 0.722 \\ 
LTH5 &   0.841 &    0.726 &  0.747 &        0.915 &        0.908 & 0.731 & 0.310 & 0.722 \\ 
LTH6 &   0.833 &    0.717 &  0.724 &        0.916 &        0.911 & 0.719 & 0.305 & 0.717 \\ 
LTH7 &   0.826 &    0.717 &  0.748 &        0.919 &        0.913 & 0.719 & 0.297 & 0.716 \\ 
LTH8 &   0.828 &    0.721 &  0.749 &        0.921 &  0.917 & 0.717 & 0.299 & 0.709 \\ \bottomrule
\end{tabular}
\end{table*}

\begin{table*}
\centering
\caption{Results using the multilayer perceptron probe, for the subset of tasks whose linear probe performances varied with sparsity.}\label{tab:mlpresults}
\begin{tabular}{lcccccccc}
\toprule
\bf{Model} &  PS-Fxn &  PS-Role &  Coref &  SynPred &  SemPred &   NER &   GED &    EF \\ 
\midrule LTH0 &   0.866 &    0.764 &  0.832 &        0.969 &        0.964 & 0.790 & 0.427 & 0.735 \\ 
LTH1 &   0.867 &    0.765 &  0.835 &        0.969 &        0.963 & 0.800 & 0.439 & 0.737 \\
LTH2 &   0.869 &    0.762 &  0.822 &        0.969 &        0.963 & 0.796 & 0.428 & 0.742 \\
LTH3 &   0.861 &    0.758 &  0.831 &        0.969 &        0.963 & 0.797 & 0.435 & 0.745 \\
LTH4 &   0.853 &    0.749 &  0.834 &        0.968 &        0.963 & 0.796 & 0.438 & 0.736 \\
LTH5 &   0.853 &    0.752 &  0.823 &        0.969 &        0.963 & 0.795 & 0.434 & 0.746 \\
LTH6 &   0.841 &    0.740 &  0.798 &        0.969 &        0.963 & 0.787 & 0.433 & 0.739 \\
LTH7 &   0.851 &    0.741 &  0.813 &        0.970 &        0.964 & 0.791 & 0.426 & 0.738 \\
LTH8 &   0.846 &    0.722 &  0.814 &        0.971 &        0.965 & 0.782 & 0.431 & 0.732 \\ \bottomrule
\end{tabular}
\end{table*}

\begin{figure*}
    \centering
    \includegraphics[width=0.8\textwidth]{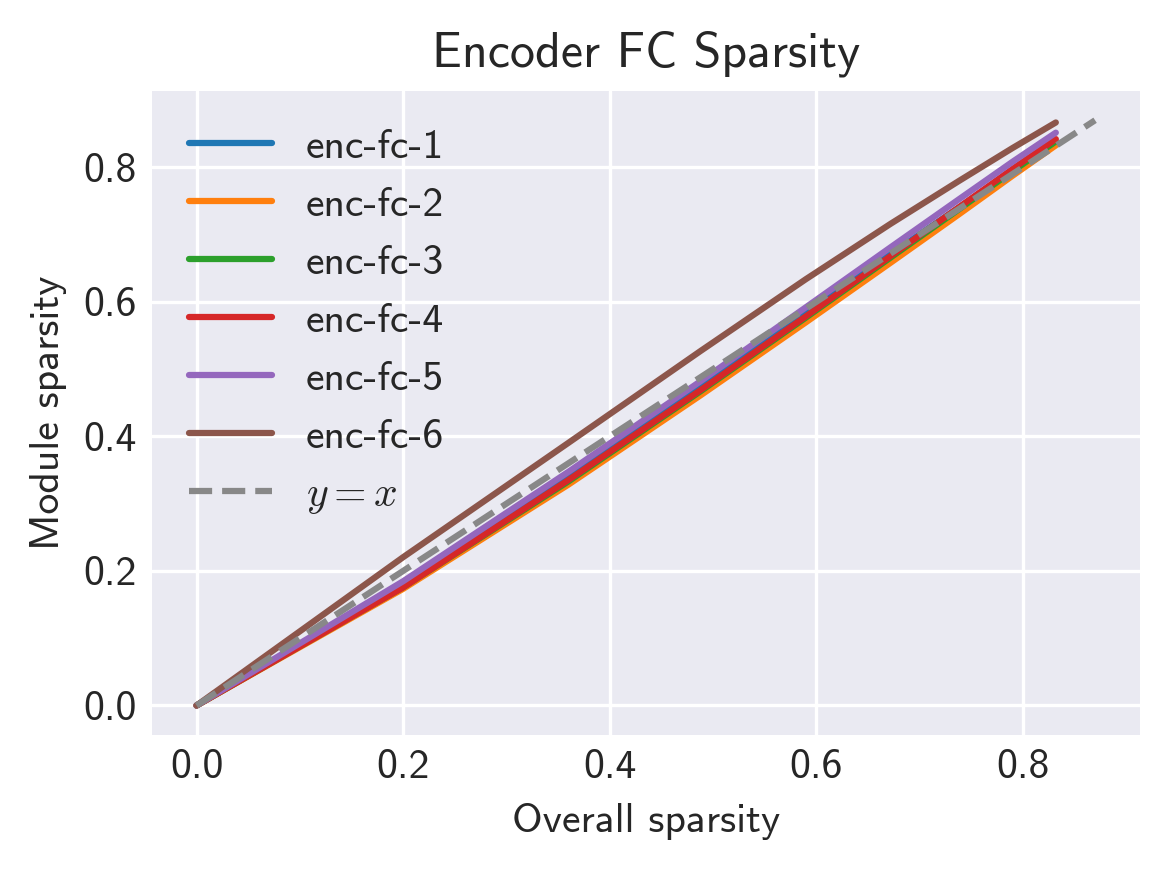}
    \includegraphics[width=0.8\textwidth]{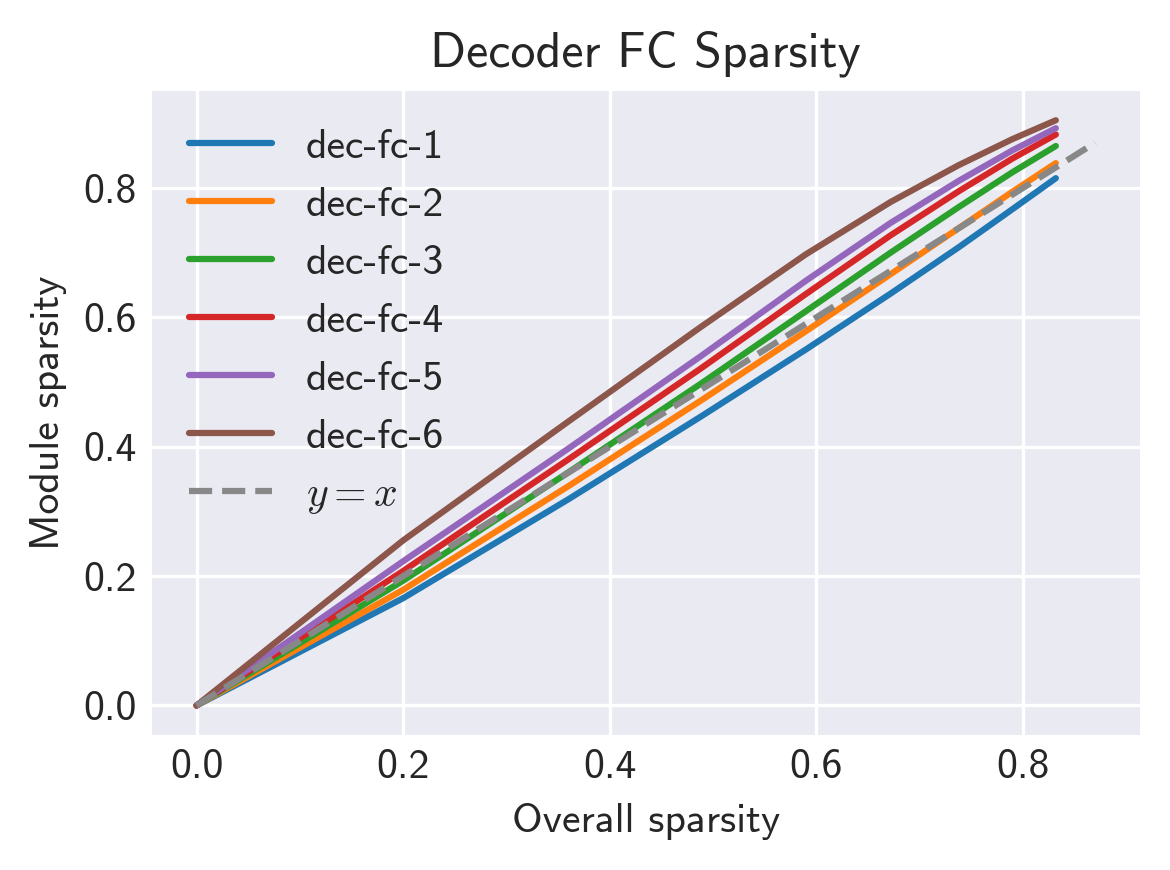}
    \caption{Sparsities of the encoder FC layers (top) and decoder FC layers (bottom). $x$-axis shows sparsity of the overall model, excluding embedding weights.}
    \label{fig:fcsparsity}
\end{figure*}

\begin{figure*}
    \centering
    \includegraphics[width=0.6\textwidth]{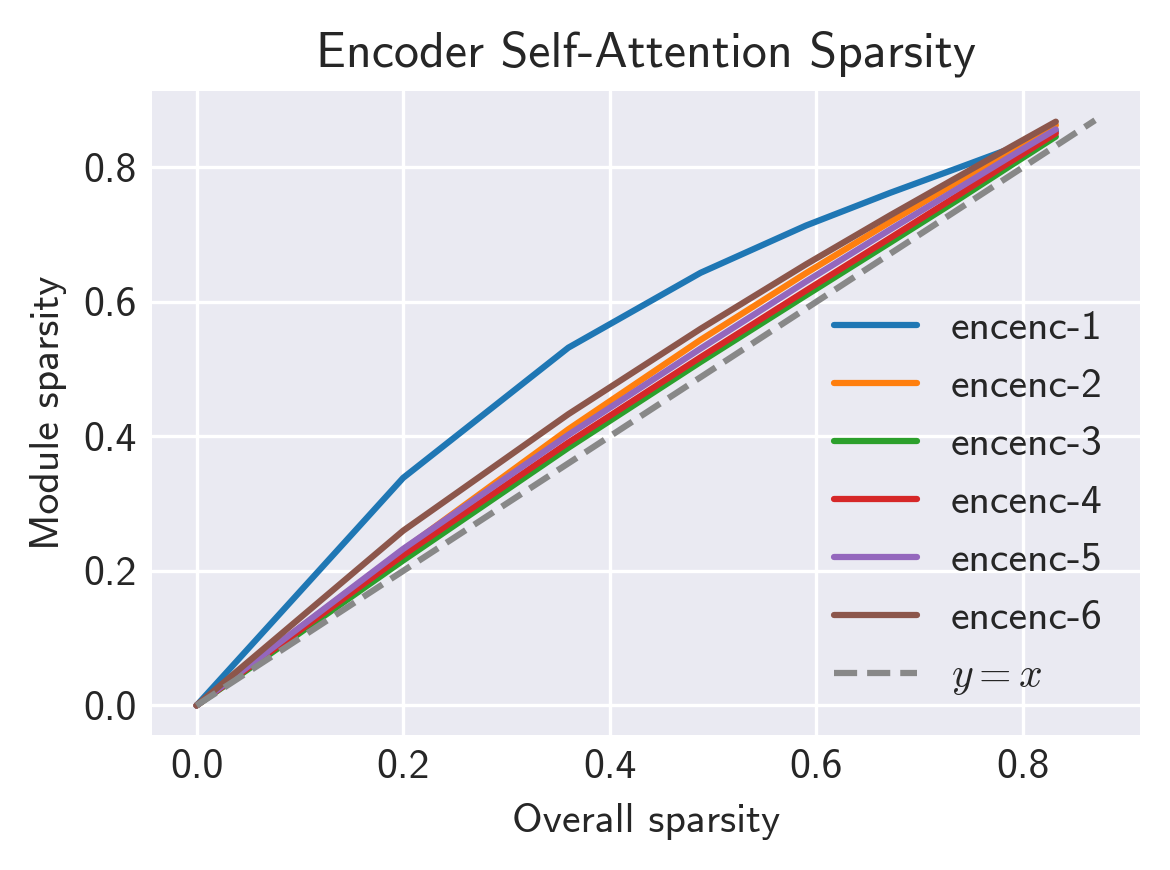}
    \includegraphics[width=0.6\textwidth]{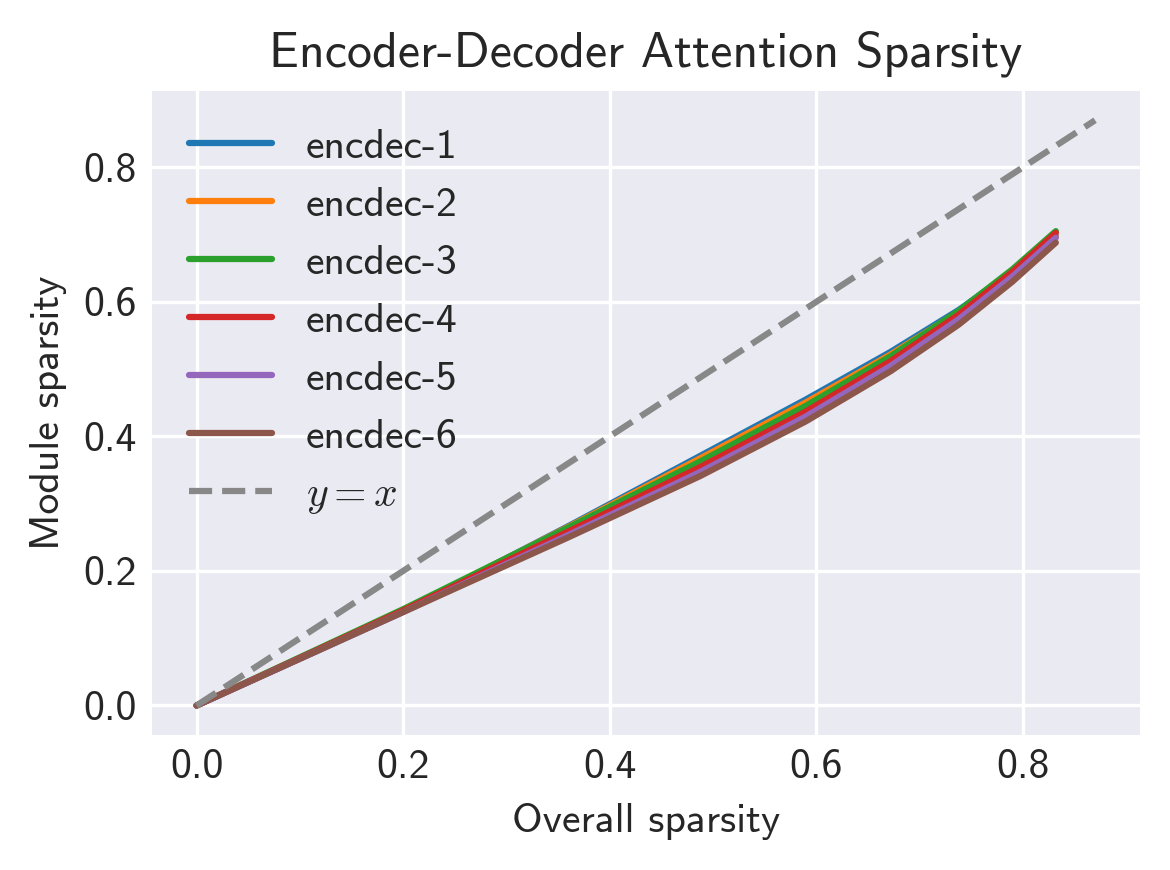}
    \includegraphics[width=0.6\textwidth]{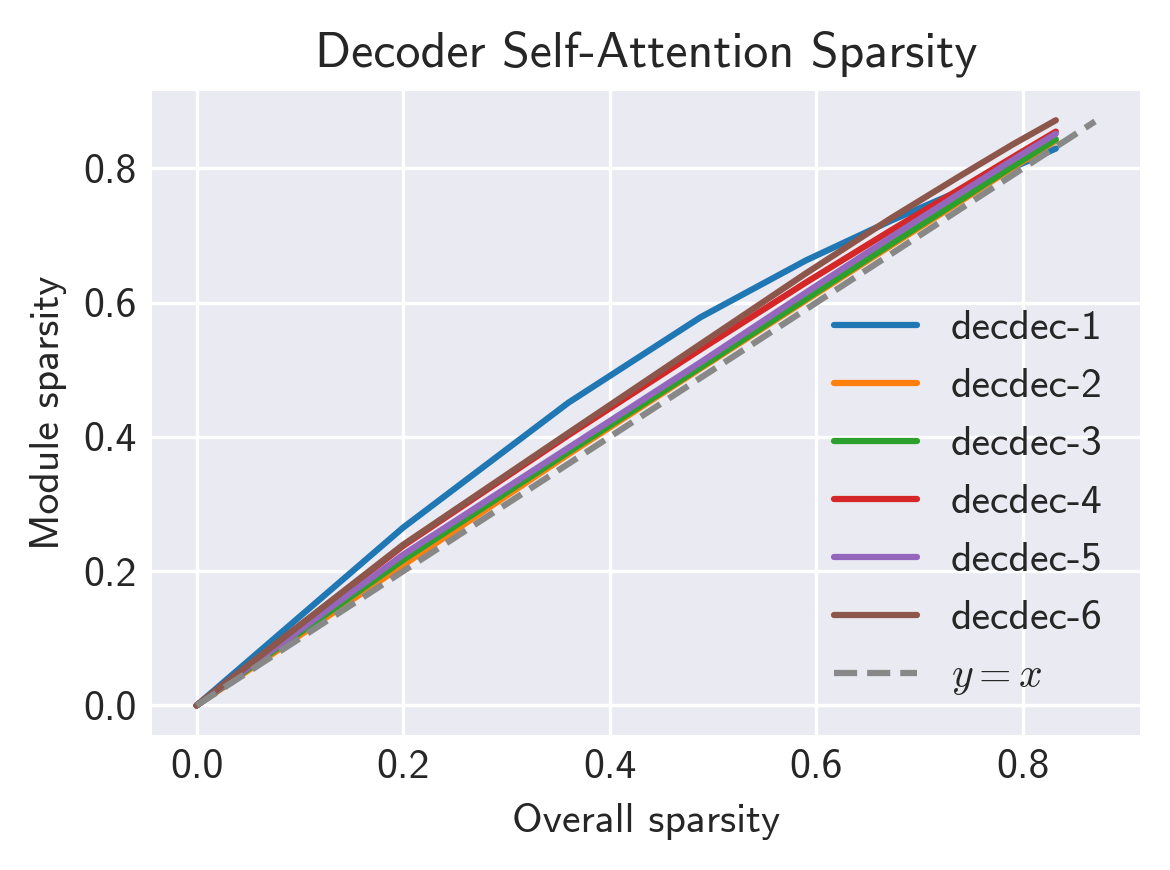}
    \caption{Sparsities of encoder self-attention (top), encoder-decoder attention (middle), and decoder self-atttention (bottom). Sparsity is aggregated across the query, key, value, and out projection matrices. $x$-axis shows sparsity of the overall model, excluding embedding weights.}
    \label{fig:attnsparsity}
\end{figure*}

\begin{figure*}
    \centering
    \includegraphics[width=0.8\textwidth]{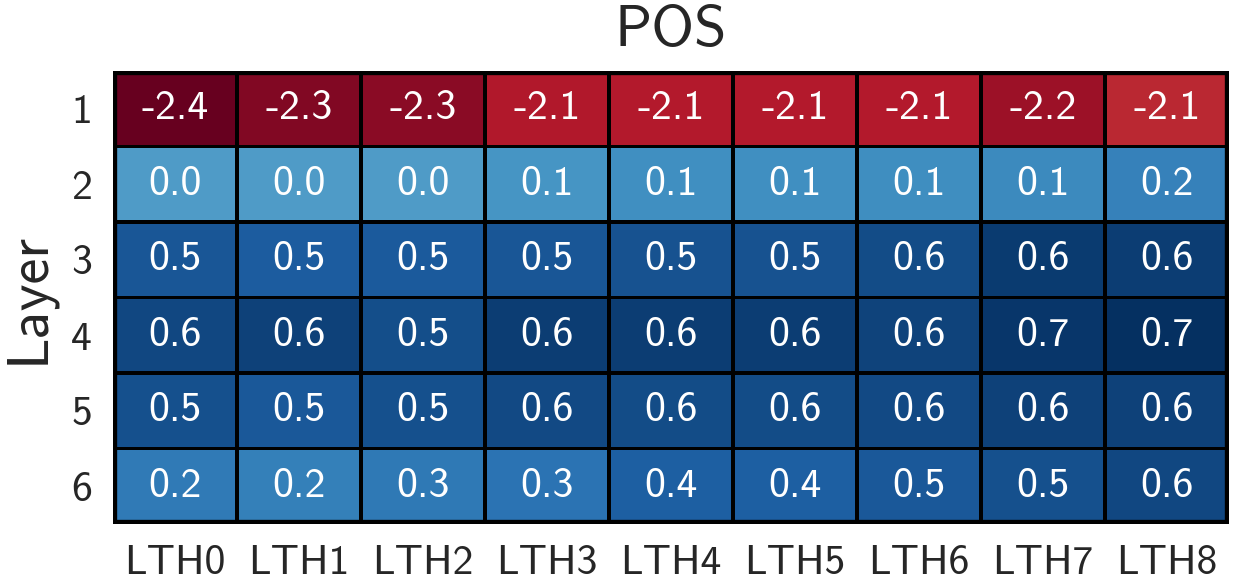}
    \caption{Each cell shows, for a particular layer of a particular model, that layer's accuracy $z$-score for the POS tagging probing task.}
    \label{fig:posprobing}
\end{figure*}

\begin{figure*}
    \centering
    \includegraphics[width=0.8\textwidth]{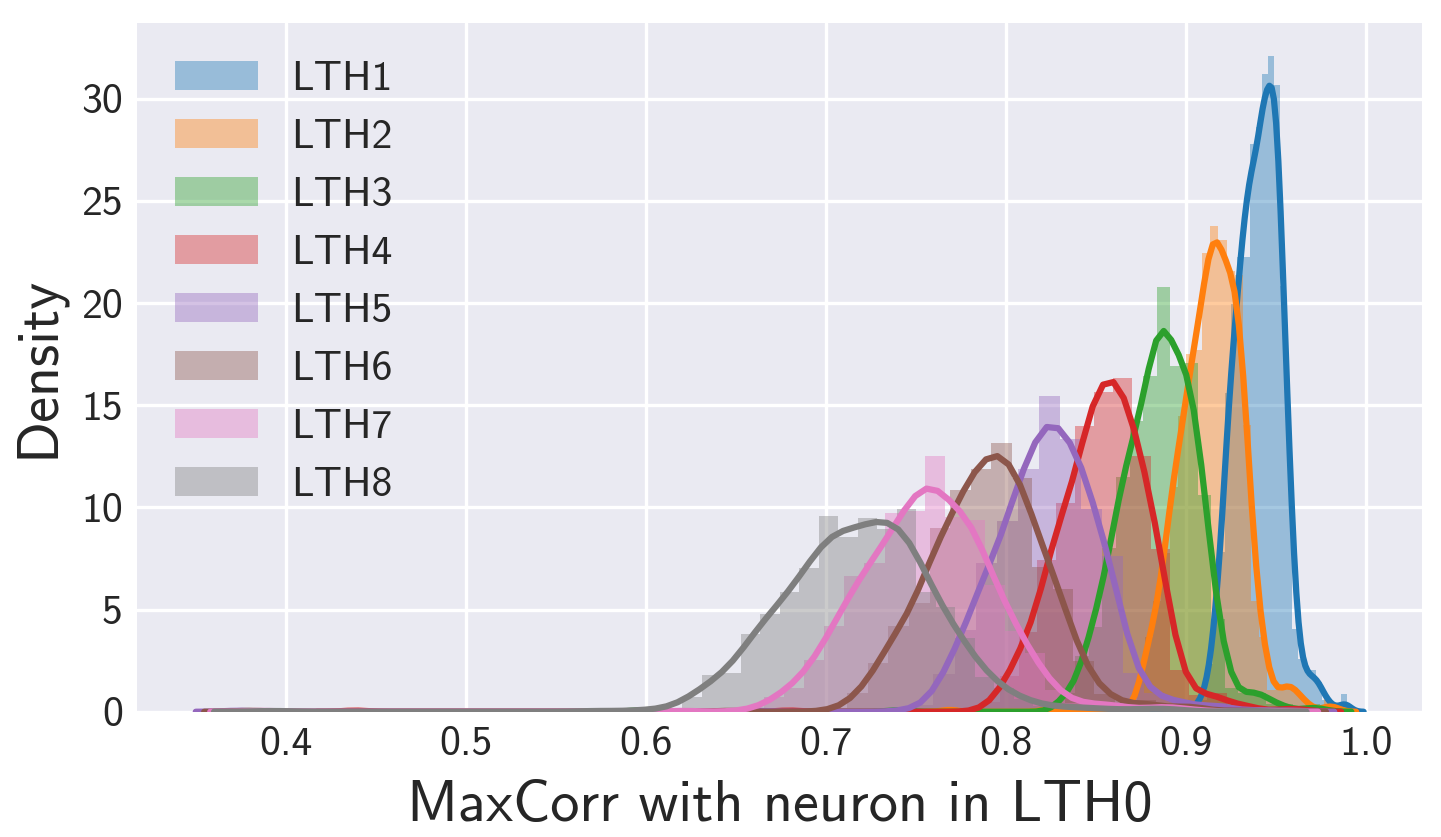}
    \caption{For each of our pruned models, we show its distribution of maximum correlations with a neuron in LTH0 (unpruned). Rather than e.g. becoming bimodal, the distributions gradually shift to the right, suggesting that all neurons slowly become less similar to their counterparts in the dense model.}
    \label{fig:maxcorrdists}
\end{figure*}

\begin{figure*}
    \centering
    \includegraphics[width=0.8\textwidth]{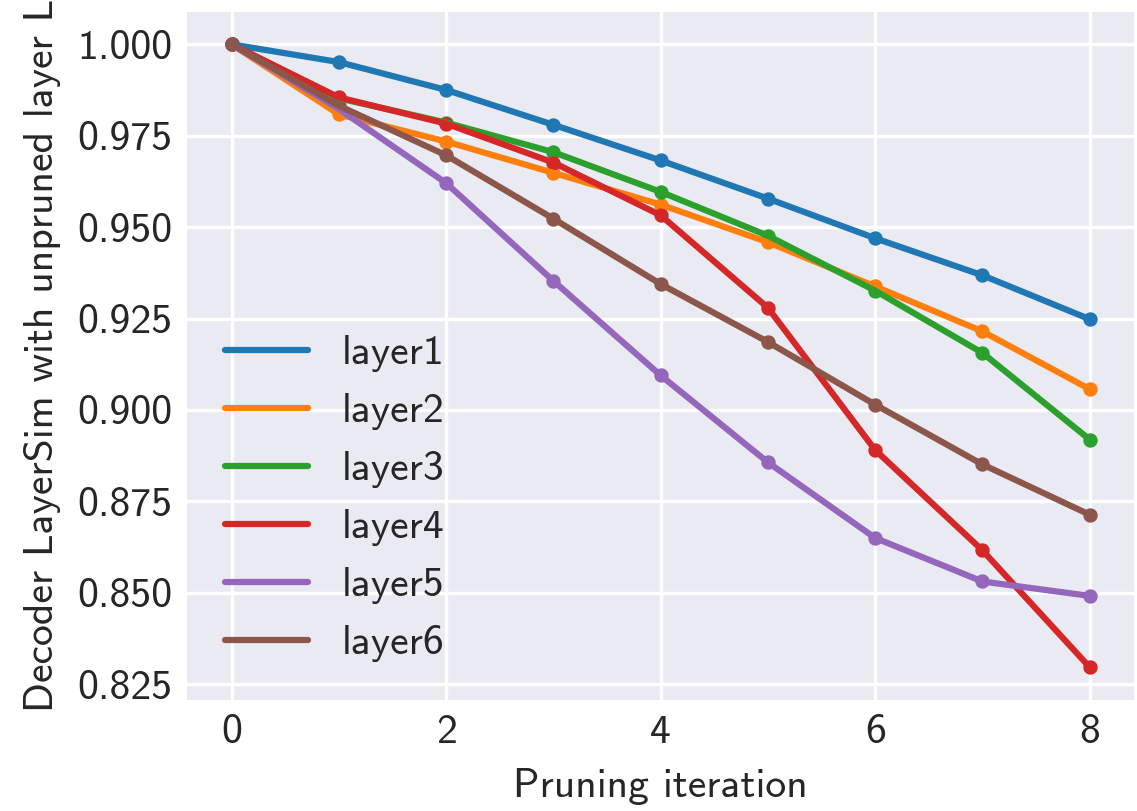}
    \caption{Each decoder layer's LayerSim with the corresponding layer in LTH0. Sparsity increases from left to right.}
    \label{fig:decoder_layersim_lineplot}
\end{figure*}

\begin{figure*}
    \centering
    \includegraphics[width=0.8\textwidth]{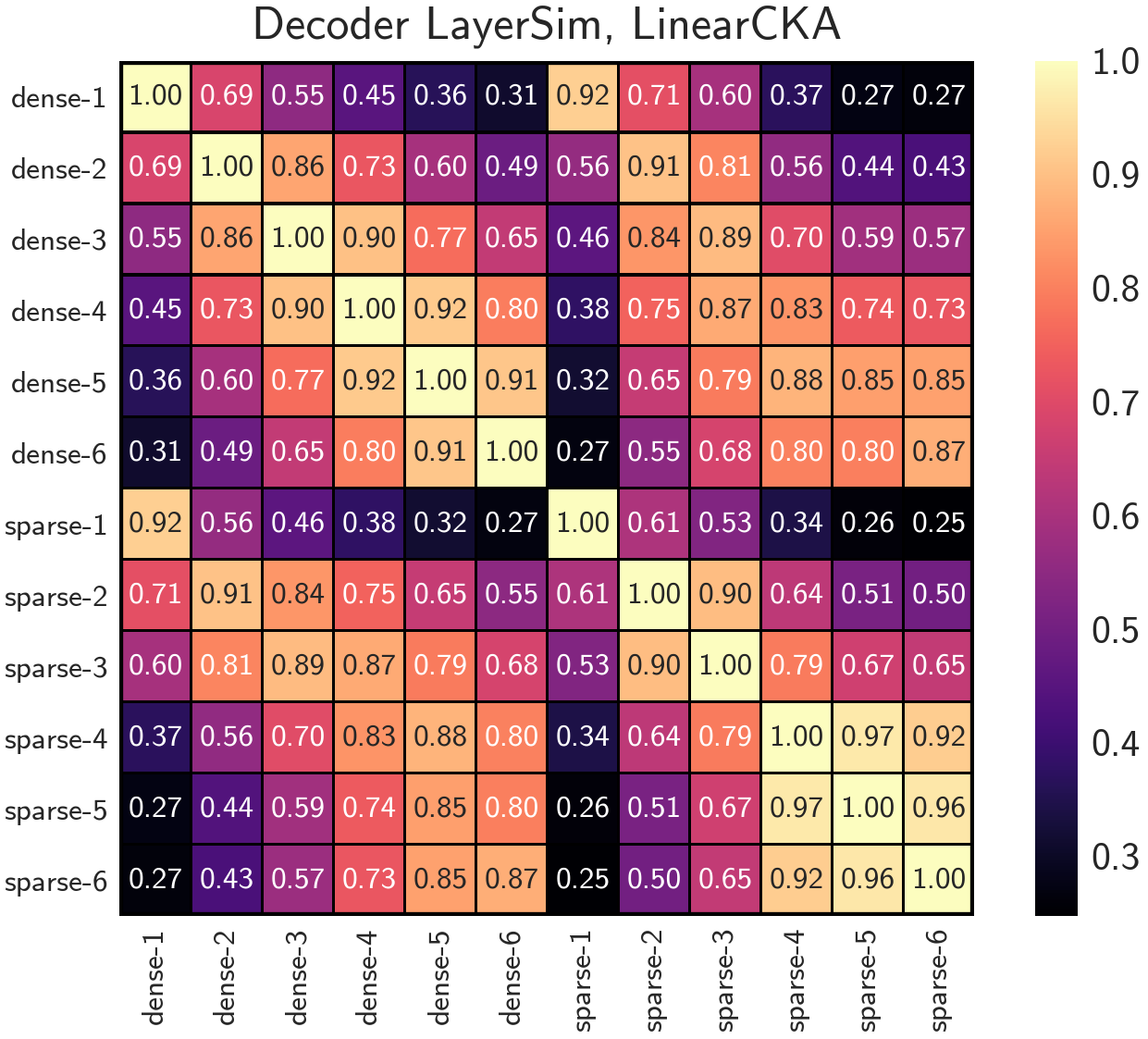}
    \caption{Decoder layer representation similarities for pairs of layers in LTH0 (dense) and LTH8 (70\% sparse).}
    \label{fig:decoder_layersim}
\end{figure*}

\begin{figure*}
    \centering
    \includegraphics[width=0.8\textwidth]{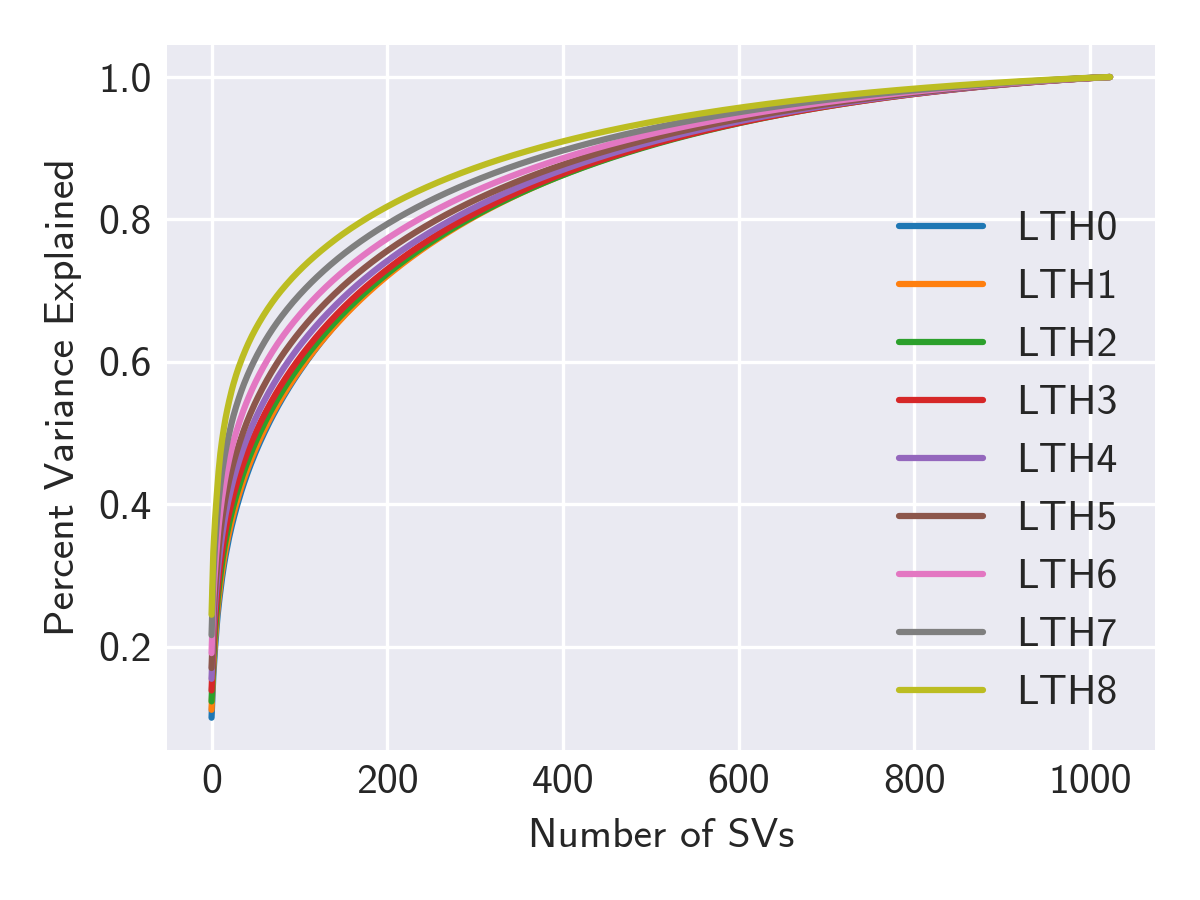}
    \includegraphics[width=0.8\textwidth]{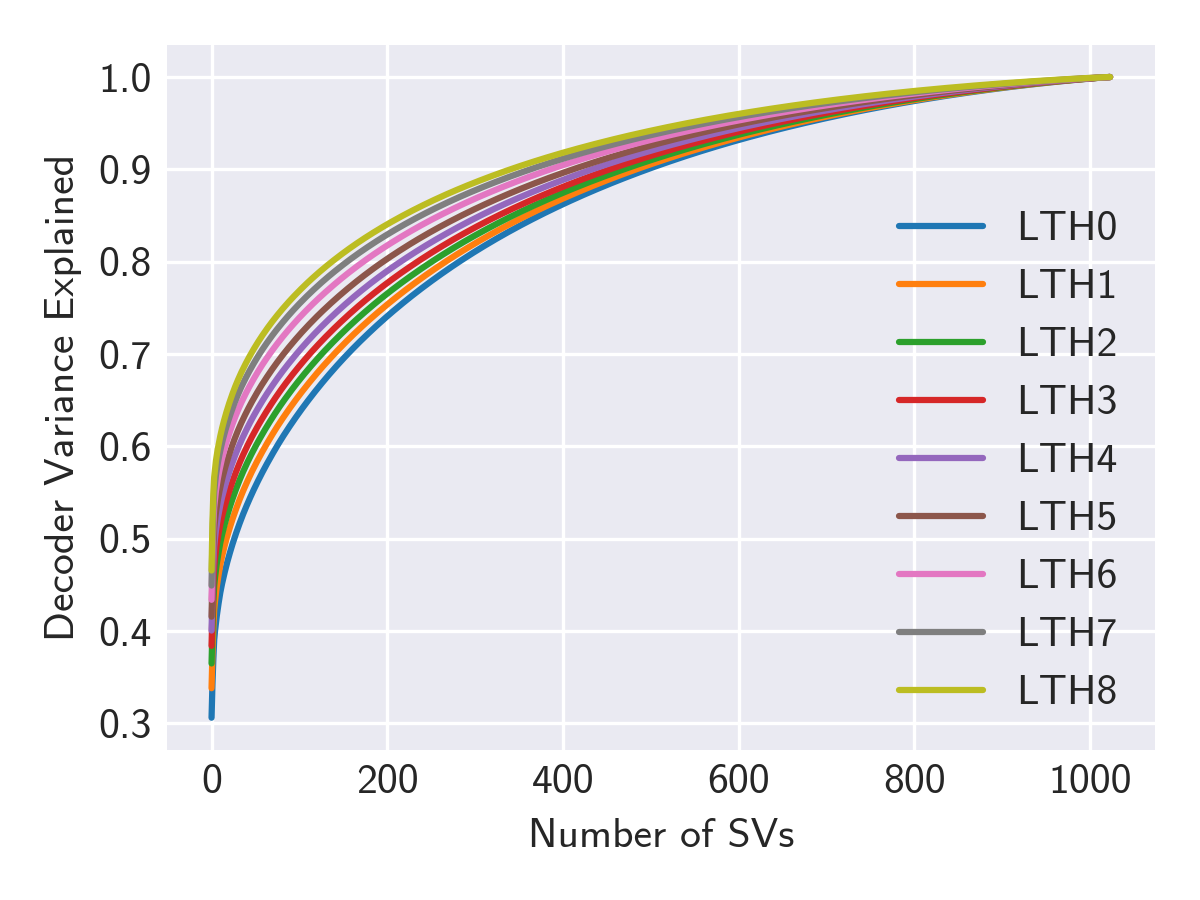}
    \caption{Percent variance explained by retaining top $k$ singular vectors in the singular value decomposition of the final layer representation matrix for the encoder (top) and the decoder (bottom). We find that sparser models have more variance explained by fewer components, implying less representational complexity.}
    \label{fig:svd}
\end{figure*}

\begin{figure*}
\centering
\includegraphics[width=0.75\textwidth]{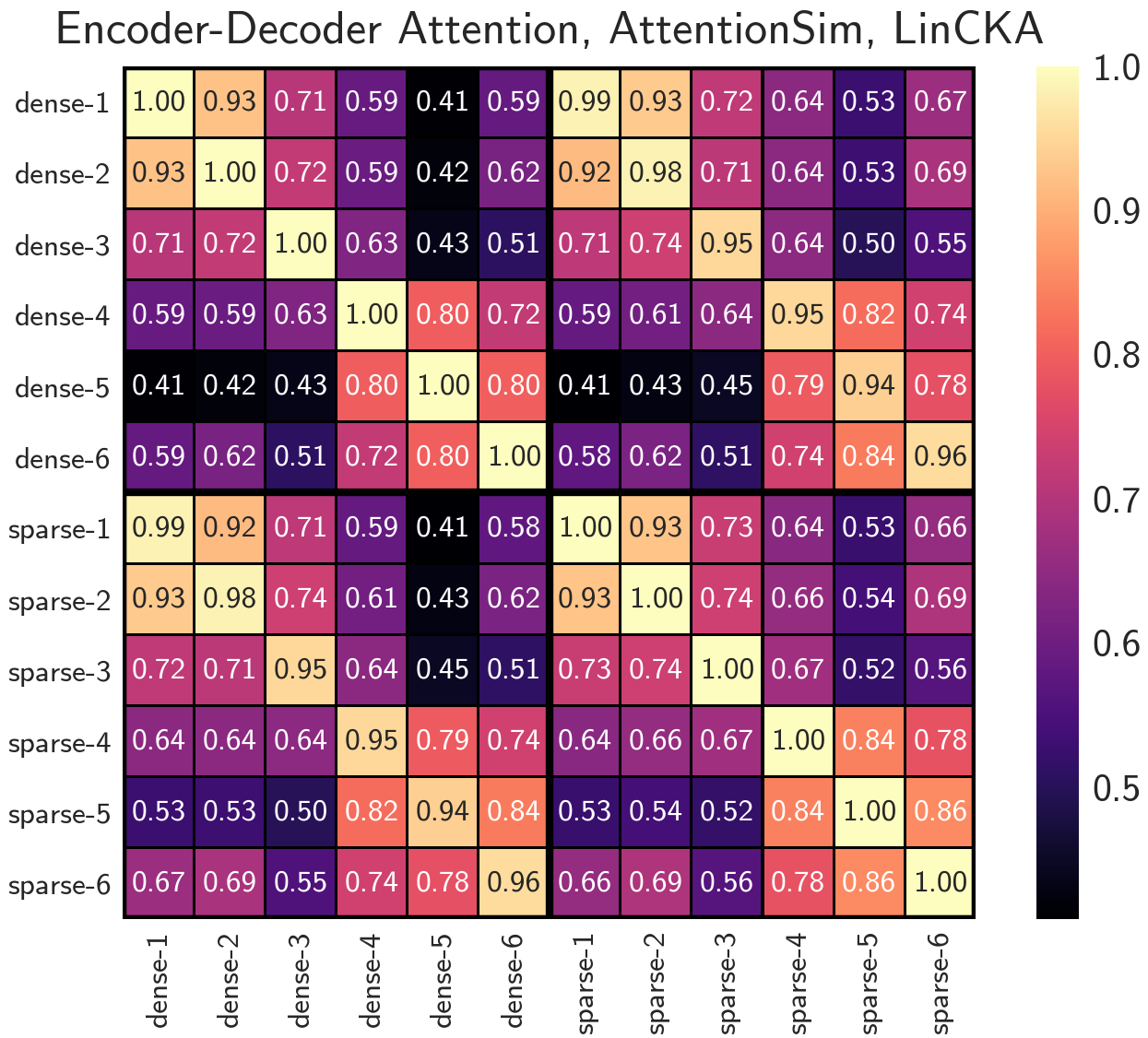}
\includegraphics[width=0.75\textwidth]{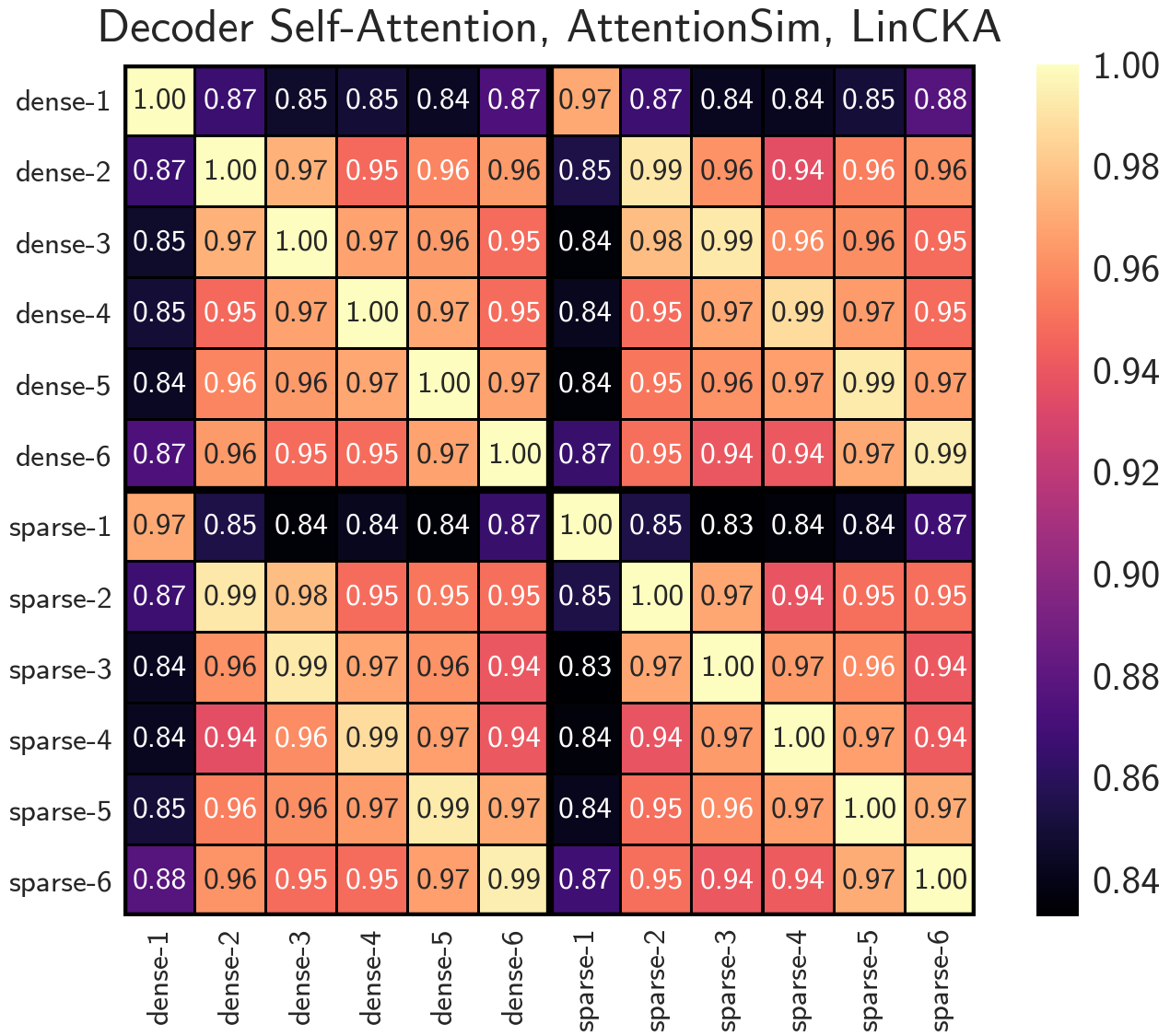}
\caption{AttentionSim similarity for pairs of layers in LTH0 (dense) and LTH8 (sparse) for encoder-decoder attention and decoder self-attention.}
\label{fig:attentionsim_small}
\end{figure*}

\begin{figure*}
    \centering
    \includegraphics[width=0.95\textwidth]{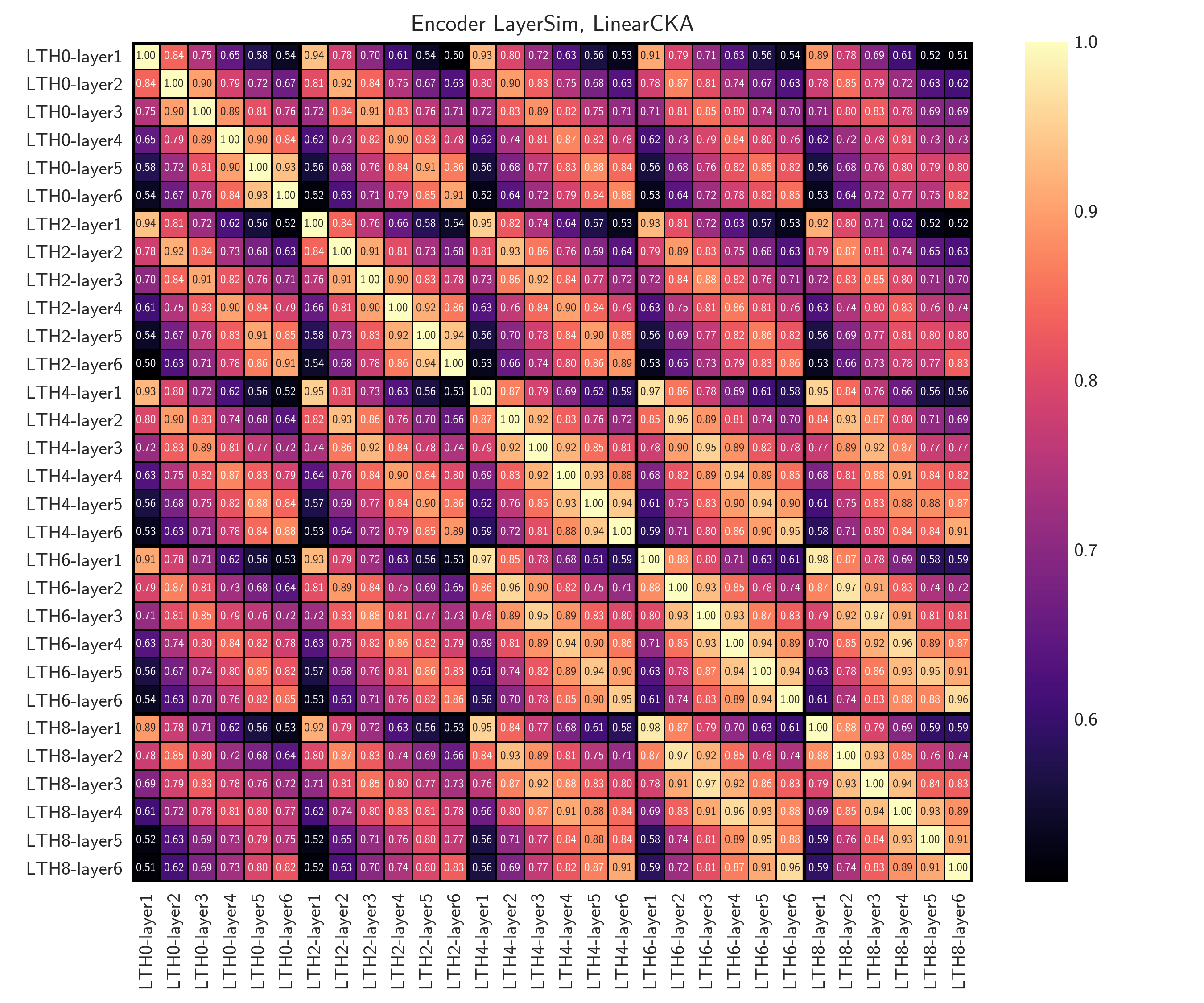}
    \includegraphics[width=0.95\textwidth]{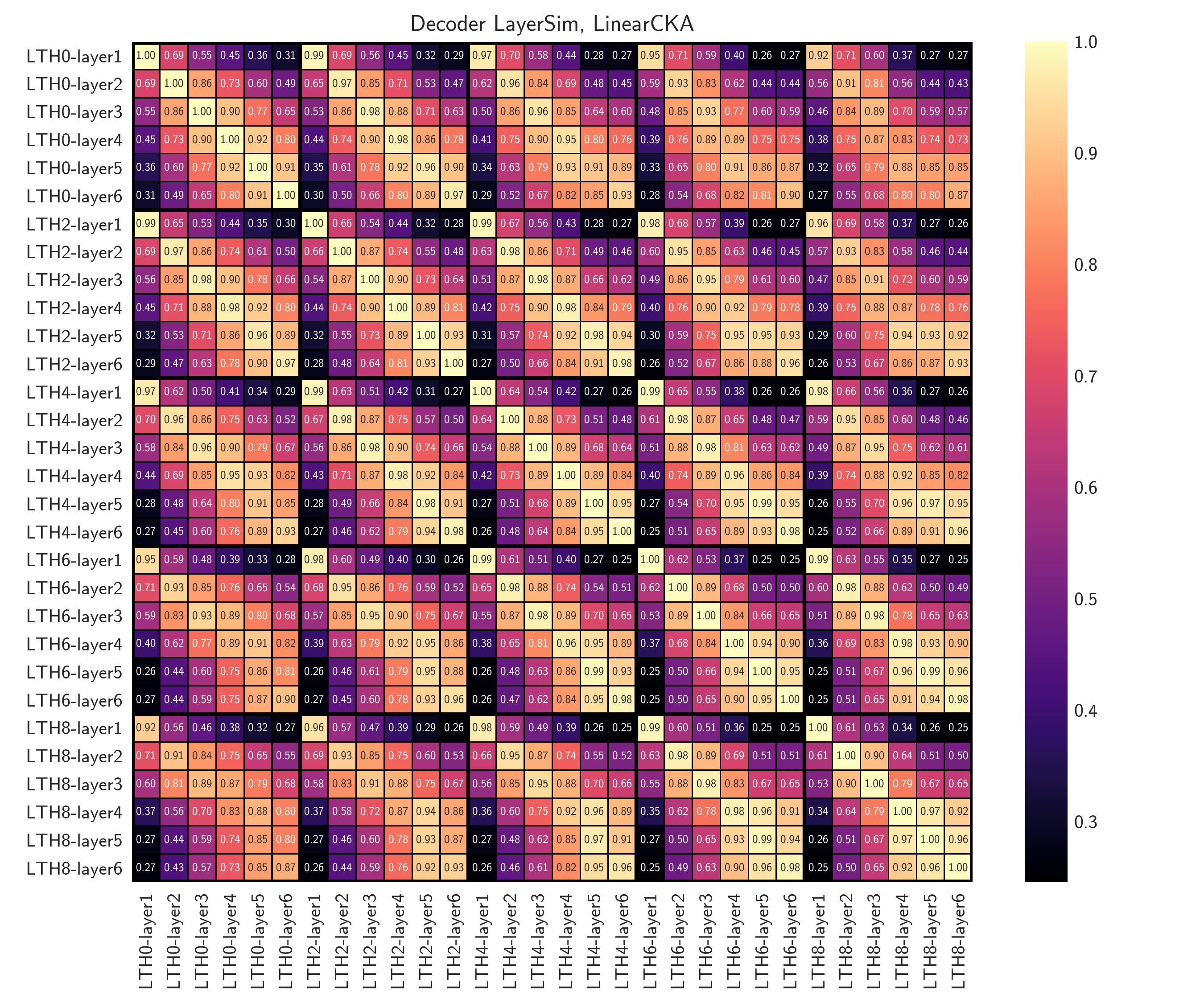}
    \caption{Full LayerSim representation similarity heatmaps for models from even-numbered pruning iterations. Top: Encoder, Bottom: Decoder.}
    \label{fig:layersim_full}
\end{figure*}

\begin{figure*}
    \centering
    \includegraphics[width=0.86\textwidth]{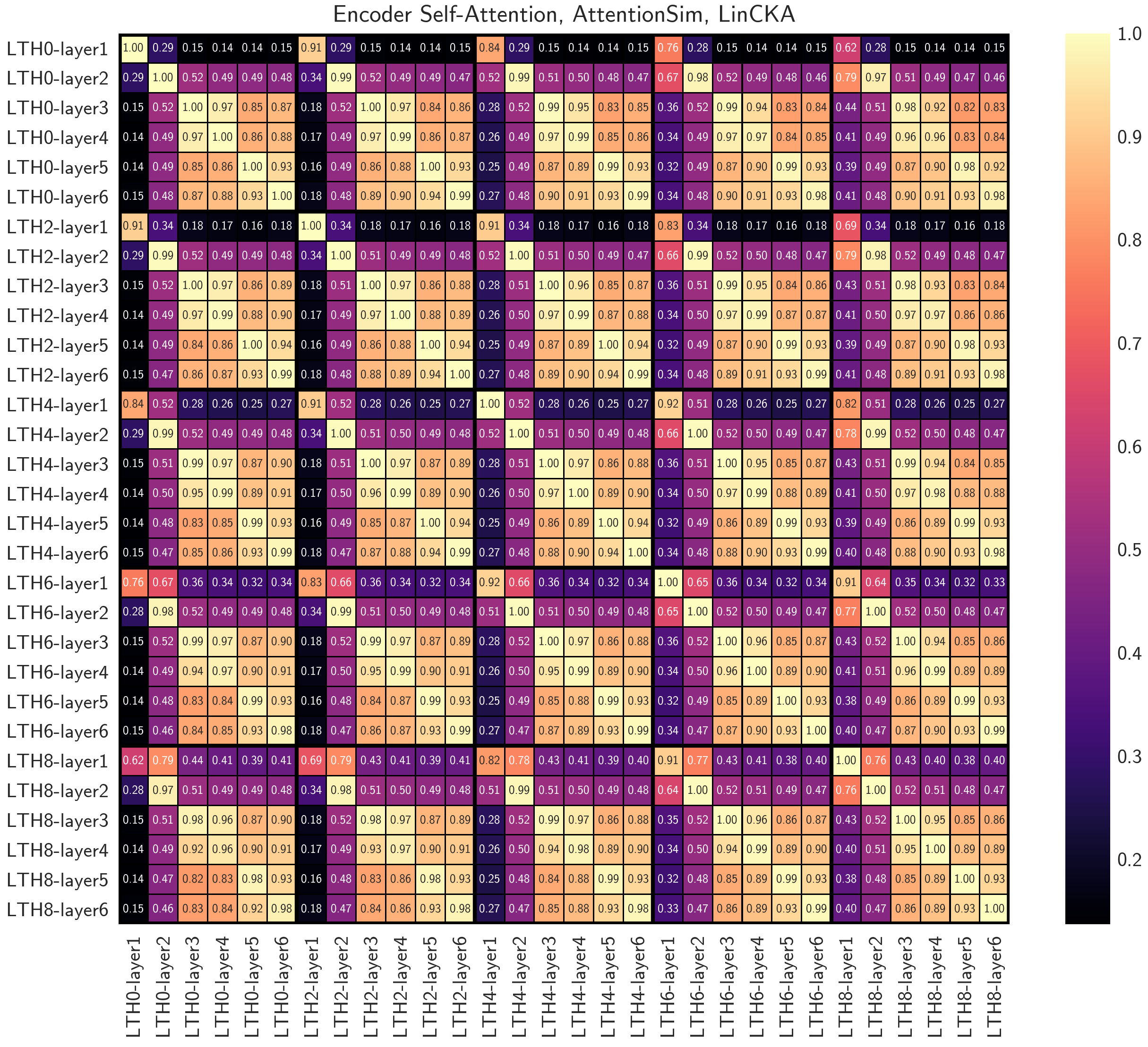}
    \includegraphics[width=0.86\textwidth]{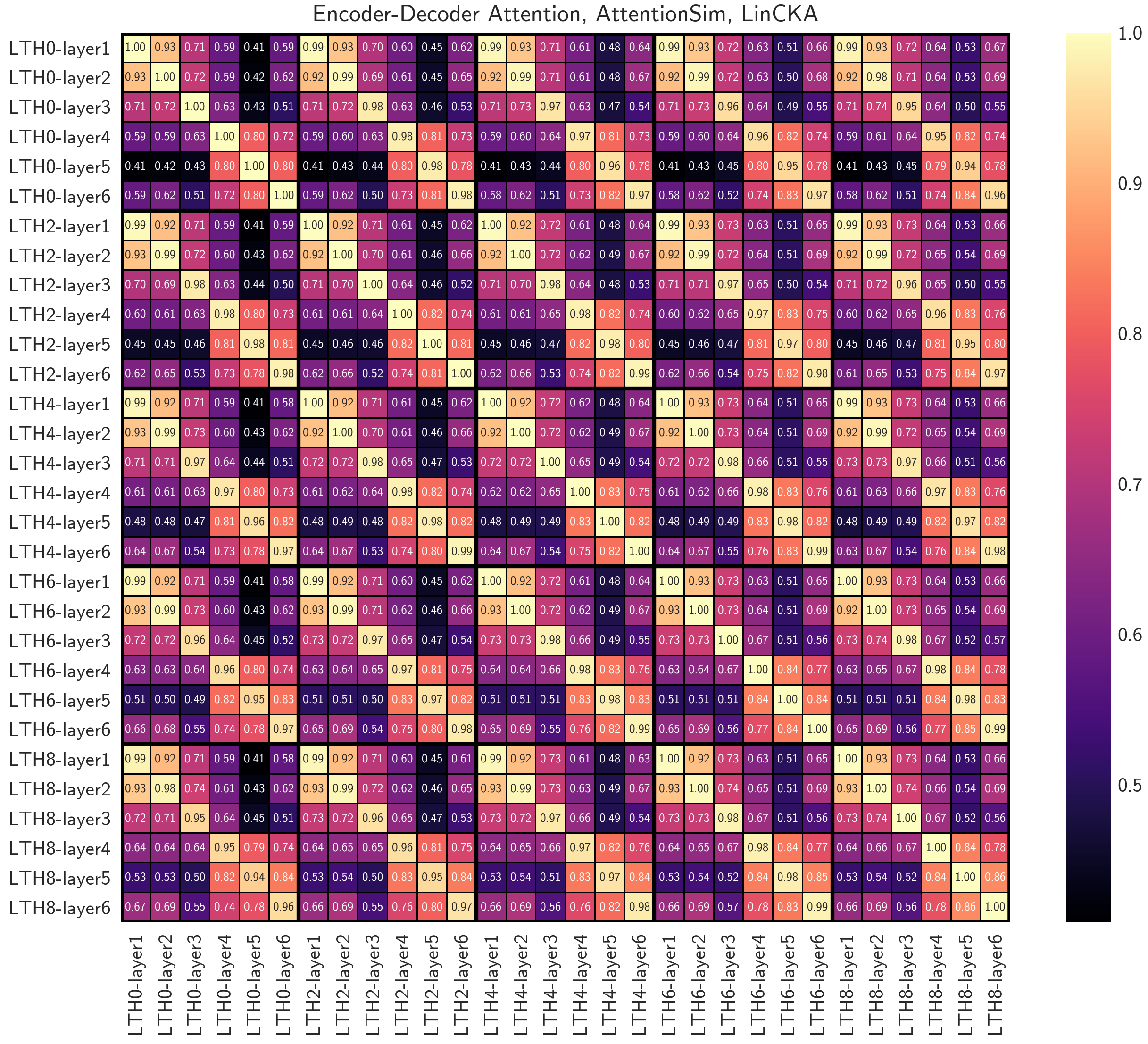}
    \caption{Full AttentionSim heatmaps for models from even-numbered pruning iterations. Top: encoder self-attention. Bottom: encoder-decoder attention.}
    \label{fig:attentionsim_full}
\end{figure*}

\begin{figure*}
    \centering
    \includegraphics[width=0.95\textwidth]{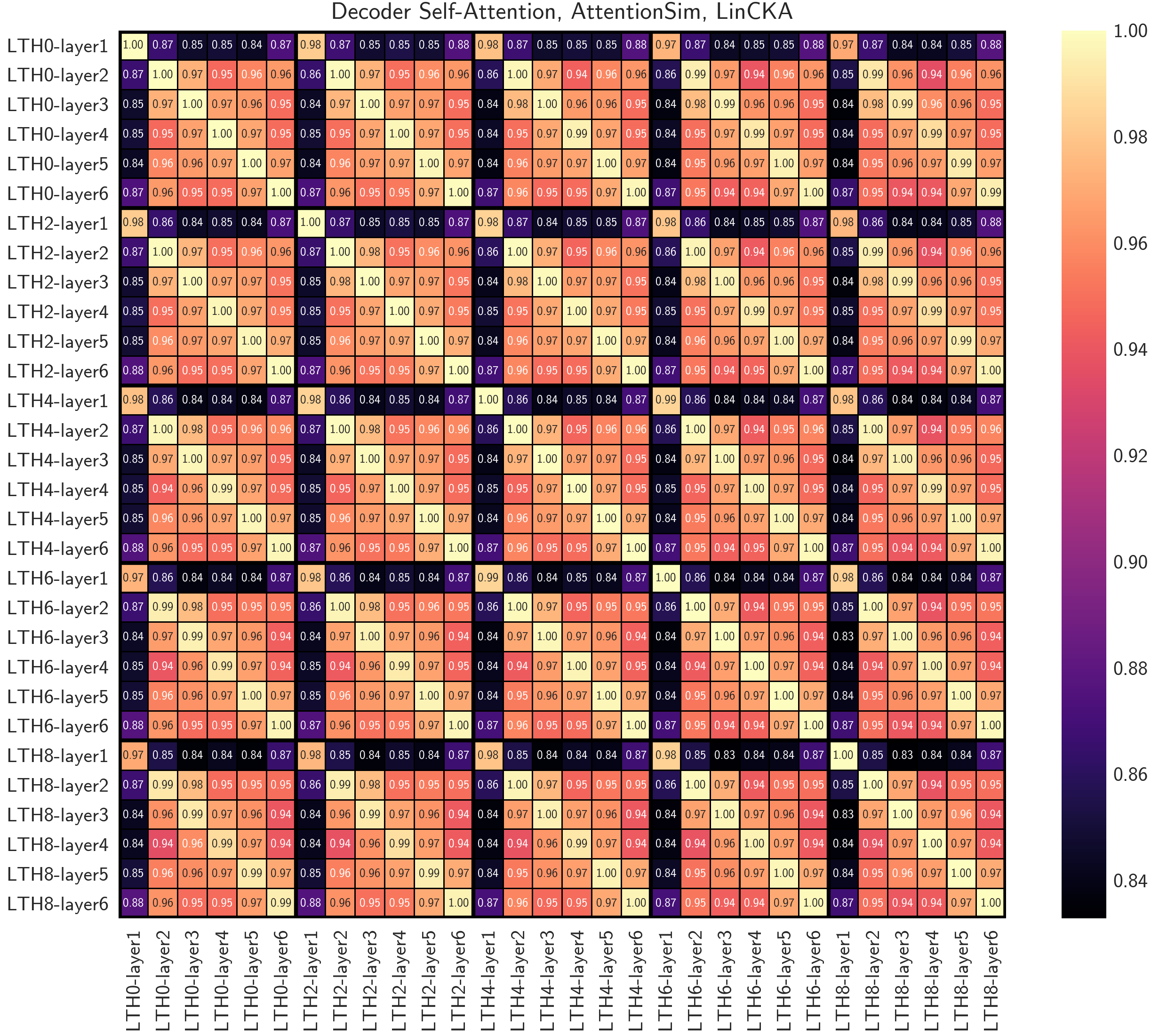}
    \caption{Decoder self-attention AttentionSim between models from even-numbered pruning iterations.}
    \label{fig:attentionsim_full2}
\end{figure*}

\end{document}